\RequirePackage{fix-cm}
\documentclass[smallextended]{svjour3}       
\smartqed  

\usepackage{graphicx}
\usepackage{xcolor} 
\usepackage{amsmath}
\usepackage{algorithm}
\usepackage{algorithmic}
\usepackage[algo2e]{algorithm2e}
\newcommand{\argmin}{\arg\!\min}
\SetAlCapSkip{1.5em}
\usepackage{algorithm2e,setspace}
\usepackage{epstopdf}
\usepackage{balance}
\usepackage[justification=centering]{caption}
\usepackage{booktabs, multicol, multirow}
\usepackage{dcolumn}
\usepackage{subfigure}
\usepackage{stackengine}
\begin{document}

\title{An Efficient Video Streaming Architecture with QoS Control for Virtual Desktop Infrastructure in Cloud Computing
}


\author{Huu-Quoc Nguyen 
	\and Tien-Dung Nguyen \and Van-Nam Pham \and Xuan-Qui Pham \and Quang-Thai Ngo 
	\and
        Eui-Nam Huh 
}


\institute{H.-Q.Nguyen 
	\and  T.-D.Nguyen \and V.-N.Pham \and X.-Q.Pham \and Q.-T.Ngo
	\and E.-N.Huh
	 \at
              Department of Computer Engineering, Kyung Hee University, 1 Seocheon, Giheung, Yongin, Gyeonggi, South Korea \\
              \email{quoc@khu.ac.kr}           
               \\
               \\
               T.-D.Nguyen\\
               \email{ntiendung@khu.ac.kr}  
                 \\
                 \\
                 V.-N.Pham\\
                 \email{nampv@khu.ac.kr}
                \\
                \\
                X.-Q.Pham\\
                \email{pxuanqui@khu.ac.kr}  
                 \\
                 \\
                 Q.-T.Ngo\\
                 \email{nqthai@khu.ac.kr} 
                        \\
                        \\
                        E.-N.Huh\\
                        \email{johnhuh@khu.ac.kr}   
}

\date{Received: date / Accepted: date}

\maketitle

\begin{abstract}
In virtual desktop infrastructure (VDI) environments, the remote display protocol has a big responsibility: to transmit video data from a data center-hosted desktop to the endpoint. The protocol must ensure a high level of client-perceived end-to-end quality of service (QoS) under heavy load conditions. Each remote display protocol works differently depending on the network and which applications are being delivered. In health care applications, doctors and nurses can use mobile devices directly to monitor patients. Moreover, the ability to  implement tasks requiring high consumption of CPU and other resources is applicable to a variety of applications including research and cloud gaming. Such computer games and complex processes will run on powerful cloud servers and the screen contents will be transmitted to the client. To enable such applications, remote display technology requires further enhancements to meet more stringent requirements on bandwidth and QoS, and to allow real-time operation. In this paper, we present an architecture including flexible QoS control to improve the users’ quality of experience (QoE). The QoS control includes linear regression modeling of historical network data as a means to consider the network condition. Additionally, the architecture includes a novel compression algorithm of 2D images, designed to guarantee the best image quality and to reduce video delay; this algorithm is based on k-means clustering and can satisfy the requirements of real-time onboard processing. Through simulations with a real-world dataset collected by the MIT Computer Science and Artificial Intelligence Lab, we represent experimental as well as explain the performance of QoS system. 
\keywords{Video streaming \and Mobile thin-client \and K-means clustering \and Quality of service \and Remote protocol \and Cloud computing}
\end{abstract}

\section{Introduction}
\label{intro}
Cloud computing has become a hot topic in todays world, largely due to the flexible and cost savings it can deliver. In the same manner as with visualization, cloud computing implementations started with server infrastructure and has moved to the desktop; the cloud is now ripe for desktop infrastructure. By moving desktops to the cloud rather than an internally deployed and managed VDI solution in a data center, businesses can realise all of the promised benefits of virtual desktops \cite{healthcare_system}. Furthermore, the Internet has rapidly emerged  as a mechanism for users to find and retrieve content, originally for webpages and recently for streaming media. Moreover, mobile phones and tablets or any thin client devices can run powerful applications from anywhere and at any time via the cloud. Therefore, there are a lot of case studies needed to transmit data from the virtual desktop in the data cloud center to the screen endpoints.

 There are several ideas for deployment in real life. In health care, while the doctors tired all the time to check a patient’s status frequently, included their heart rate or other medical signs read by specialized equipment. With VDI solutions, a patient can get convenient health care from the comfort of their own home. This allows doctors to quickly and conveniently make such information available remotely on the mobile devices which are connected to a server by using a remote desktop protocol \cite{healthcare_system}. Moreover, VDI helps save valuable time and satisfies the demand for personal control, while also reducing the cost of long-term medical care by avoiding waste of human resources on manual processing \cite{personal_cost}. Because any mistakes is always leading cause of death, so new developments in health care are of considerable interest. Computer gaming is a very different application where remote display technology will also be useful. Gamers spent a lot of money on computer games, hardware, and accessories. Traditionally, computer games are delivered either in boxes or via Internet downloads. These games have to install the computer games on physical machines to play them. The installation process becomes extremely tedious because the games are too complicated and the computer hardware and system software are very fragmented. Take Blizzard’s Star-craft II as example, it may take more than an hour to install it on an i5 PC, and another hour to apply the online patches. Furthermore, gamers may find their computers are not powerful enough to enable all the visual effects yet achieve high frame rates. So, gamers have to repeatedly upgrade their computers so as to play the latest computer games. Hence, cloud gaming is a better way to deliver high-quality gaming experience and opens new business opportunity \cite{cloud_gaming}. In a cloud gaming system, computer games run on powerful cloud servers and streaming video data to thin clients by using remote protocols \cite{thinclient_gaming}. The thin clients are light-weight and can be ported to resource-constrained platforms, such as mobile devices and TV set-top boxes. With cloud gaming, the key benefit is that gamers can play the latest computer games anywhere and anytime, while the game developers can optimize their games for a specific PC configuration. Laboratory research is another excellent potential application for cloud-based remote display technology. By using remote desktop protocols, researchers perform complex tasks requiring intensive computation resources or specialized software. In addition to these example applications, there are other fields in which using remote display protocols would be useful for transmitting screen data from a cloud server to a client.

\begin{figure*}
	\includegraphics[width=0.75\textwidth]{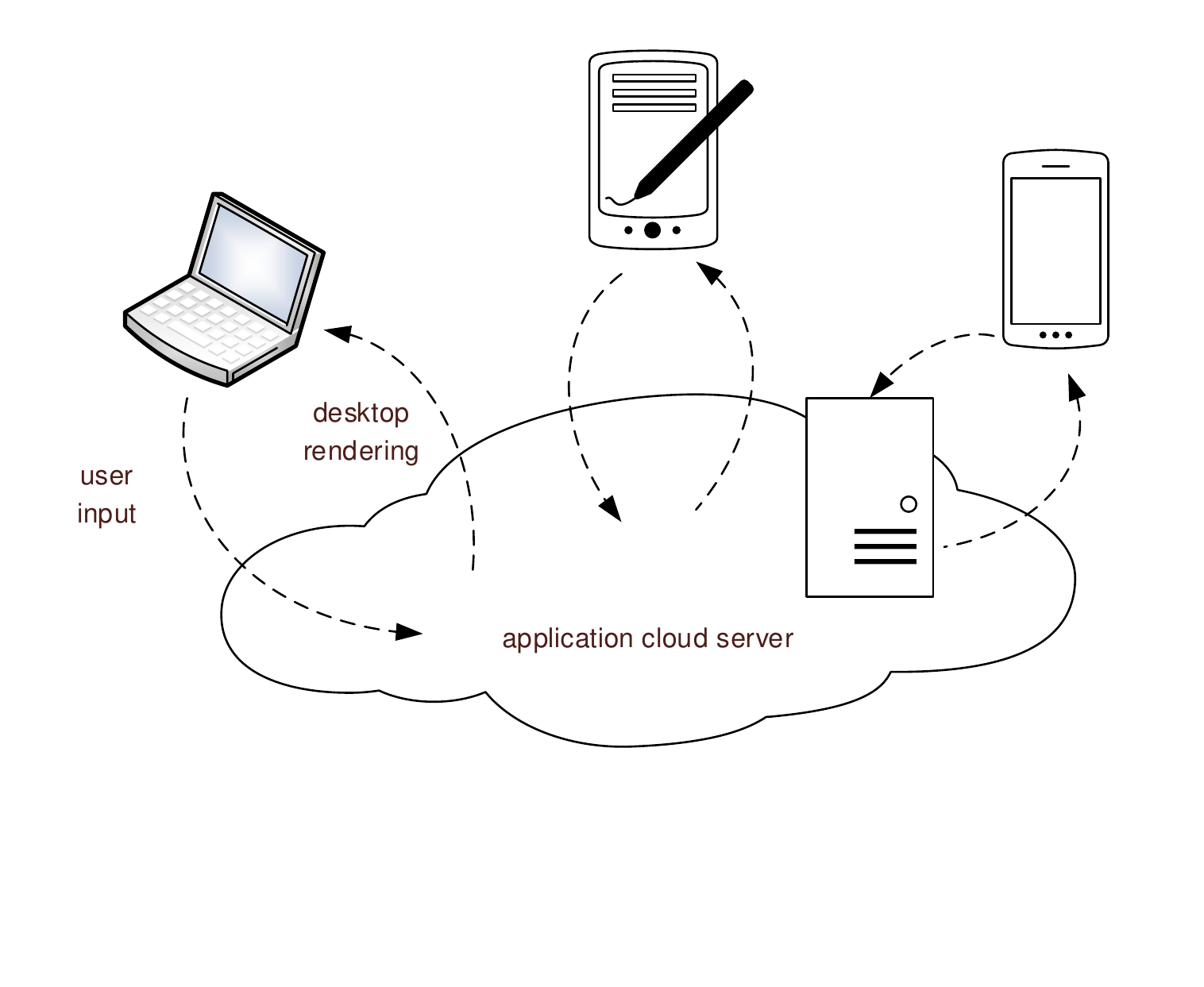}
	\centering
	\vspace{-1em}
	\vspace{-3.75em}
	\caption{General model of desktop rendering between client and cloud server}
	\label{fig_client_server}       
\end{figure*}

 However, in the past there have been several challenges with deploying VDI. Main problems included the network condition, server overload, bandwidth and limitation of mobile device (hardware resource, storage, battery life time). Specifically, a client sends an input event request to a server, the sever receives and deploys several offloading computing tasks, then renders screen image, encodes, and finally transfers the results to the client (as shown in Fig~\ref{fig_client_server}). These graphical images will be decoded and displayed on the client side. Therefore, it is so importance to guarantee client-perceived end-to-end quality of service. We can summarize several important factors to measure performance of the system:
\begin{itemize}
	\item Network condition: latency and bandwidth utilization. Latency should be as small as possible. Furthermore, with many different bandwidth quotas, users always want to get the highest acceptable performance. 
	\item Hardware resource consumption: CPU has a critical meaning in the processing because the limited hardware resource of the client device and overhead consumption on the server. 
	\item User demands: Cross-platform adaptation because it is a great convenience for users who use many kinds of devices such as laptop, smart phone, tablet, etc. with different operating systems.  
\end{itemize}

Above factors directly affect QoE (Quality of Experience) that user perceives in each request which is made from local desktop, where claims for more high fidelity of display. Many solutions have been developed and reported for improving QoE in content delivery networks; typically, solutions include hardware and software improvements. The general idea is that whatever graphical elements are painted to the screen on the host and then scraped by the protocol interface, are sent down to the client. This can happen in two ways:
\begin{itemize}
	\item The client can contact the server and pull a new snapshot of the screen from the frame buffer. This is how Virtual network computing (VNC) works \cite{vnc1}, \cite{vnc2}. 
	\item The server can continuously push its screen activity to the client. This can be at the frame buffer level, the GDI / window manager level, or a combination of both. This is how (Remote Desktop Protocol) RDP \cite{rdp} and Independent Computing Architecture work \cite{ica}.
\end{itemize}

In this paper, to avoid these above problems we present an architecture with flexible QoS control based on network condition analysis to improve users’ QoE. We carefully establish the best compression parameters based on statistical analysis of historical data on network conditions. Moreover, to improve the real-time video streaming quality we develop a novel compression algorithm for 2D images, based on k-means clustering. The contributions of this work can be summarized as follows:

\begin{enumerate}
	\item We develop a novel image compression algorithm based on k-means clustering which can be applied in real-world case studies.
	\item We present an efficient architecture including QoS control based on analysis of historical data to account for network conditions. This system applies multiple linear regression to make optimal decisions for the QoS policy. 
\end{enumerate}
We evaluate the proposed image compression algorithm by means of simulations in Matlab, using as input a dataset of real images collected by the MIT Computer Science and Artificial Intelligence Lab \cite{image_dataset}. We demonstrate the QoS control system in a testbed cloud environment and analyze its performance.

The rest of this paper is organized as follows. Related work is reviewed in Section 2. Section 3 gives an image compression algorithm based on the k-means algorithm. Section 4 suggests an architecture to dynamically modify the compression rate of a video stream based on historical data of the recipients bandwidth. Section 5 describes experiments with the proposed QoS control and evaluates its performance. Finally, Section 6 gives conclusions.

\section{Related Work}
Popular remote display protocols offer high-resolution sessions, multimedia stream remoting, multi-monitor support, dynamic object compression, USB redirection, drive mapping and more. Microsoft Remote Desktop Protocol (RDP), VMwares PC-over-IP (PCoIP) and Citrixs HDX are the most com- monly used, but there are other protocols from companies such as Ericom and Hewlett-Packard. Teradici sells hardware solutions based on PCoIP technology \cite{pcoip} and the Thin Client Internet Computing protocol is also applicable for remote displays \cite{thinc}. But these protocols still have drawbacks; great performance requires a lot of bandwidth and energy, which can clog the network \cite{energy_solution}. Plus, if the system is running with low CPU usage, protocols will hog bandwidth and weaken end user performance. So, an efficient system needs to cope with unstable network conditions, to operate in real time, and to satisfy the users’ demands. Almost all of these protocols have the disadvantages that they do not consider the network environment and are inconvenient to use with real-time applications such as video playback. Previous work has included the development of an adaptive encoding application to solve display performance problems encountered when using RDP \cite{related1}. The authors of another work have described problems that arise in the case of video conferencing and have developed a real-time camera control system to address the problems \cite{related2}. 

To reduce the size of images transferred over the network to improve video streaming quality, compression technology has been one of the most important contributions. Some encoding and decoding formats such as motion MPEG \cite{related3}, \cite{related5} and H.264 AVC \cite{related4}, \cite{related6} have been developed in hybrid remote display protocol. The challenges for compressing images have been summarized previously \cite{related7}. Various compression techniques have been developed for image compression \cite{image_compression1}, \cite{image_compression2}.  

In this work, we focus on the k-means clustering for encoding the image data. This method  is probably the most popular technique of representative-based clustering. Cluster analysis \cite{clustering} is a major data analysis method that is widely used for many practical applications in emerging areas. Clustering is the process of partitioning a given set of objects into disjoint clusters. This is done in such a way that objects in the same cluster are similar whereas objects belonging to different clusters differ considerably with respect to their attributes \cite{k_means_}. The k-means algorithm \cite{kmeans1}, \cite{kmeans2} is effective in producing clusters for many practical applications. But the original k-means algorithm is very computationally complex, especially for large data sets. Moreover, this algorithm yields different clusters depending on the random choice of initial centroids. For example, previous work has included the description of a 2D image compression scheme using k-means in a real-time multiprocessor system \cite{kmean_compression1}. Another work presented a novel k-means algorithm for compressing images, including exploration of the strategy of applying statistical parameters for choosing the initial seeds \cite{kmean_compression2}. Despite the efficiency of the latter scheme, it still has drawbacks like a priori fixation of the number of clusters and random selection of initial seeds. Various researchers have made several attempts to improve the performance of k-means clustering \cite{kmean_compression3}, \cite{kmean_compression4}, \cite{kmean_compression5}. In addition, hierarchical clustering algorithms also usually apply in image processing. But as the number of records increase the performance of hierarchical algorithm goes decreasing and time for execution increased.
K-mean algorithm also increases its time of execution but as compared to hierarchical algorithm its performance is better. As a general conclusion, k-means algorithm is good for large dataset and hierarchical is good for small datasets \cite{compare_kmeans_hierarchical}. Therefore k-means is most suitable for image processing in the video streaming scheme.

Beside, developing a model that efficiently provides QoS control is presently a challenging problem. Many researches have addressed the problem of providing QoS guarantees to various network applications and clients. Most of them have focused on servers without considering network delays \cite{related8}, \cite{related9}, on individual network routers \cite{related18}, \cite{related19}, \cite{related20}, or on clients with assumptions of QoS supports in networks \cite{related21}. Some recent work focused on end-to-end QoS guarantees in network cores \cite{related22}, \cite{related23}, with the aim of guaranteeing QoS as measured from server-side network edges to client-side network edges. 

To solve this problem, previous approaches need to be revised to allow their application to a cloud environment in a manner that responds to various user behaviors and network conditions. In the present work, k-means clustering is used to reducing the size of one frame from the video. We then develop a scheme for the analysis of historical data as a means to consider network status and make a decision regarding the optimal compression ratio for QoS control. Eventually, we evaluate our system model using actual measurement results resulting from the processing of images from a realistic image dataset.

\section{Image Compression Module}
Clustering methods are widely used in pattern recognition, data compression, data mining, but the problem of using them in real-time systems has not been a focus of most algorithm designers. In this paper, we describe a practical clustering procedure that is designed specifically for compression of 2D images.
\subsection{K-means Clustering}
The k-means algorithm is one of the most commonly used methods for data clustering \cite{k_means_algorithm}. Given a data set	$X = \{x_{1}, x_{2},\dots, x_{N}\} \in {\rm I\!R}^D$ the objective of k-means is to partition X into K exhaustive and mutually exclusive clusters $S = \{S_{1}, S_{2},\dots, S_{K}\}, \bigcup_{k=1}^KS_k = X, S_i \cap S_j = \emptyset$ for $1 \le i \neq j \le K$ by minimizing the sum of squared error (SSE):
\begin{equation}\label{Eq1}
{SSE = \displaystyle\sum_{k=1}^{K} \displaystyle\sum_{x_i \in S_k} \begin{Vmatrix} x_i - c_k\end{Vmatrix}_2^2}
\end{equation}
where $\begin{Vmatrix}\  \end{Vmatrix}_2$ denotes the Euclidean norm \cite{euclidean} and $c_k$ is the centroid of cluster $S_k$ calculated as the mean of the points that belong to this cluster. This problem is known to be NP-hard even for K = 2 \cite{np1} or D = 2 \cite{np2} because clustering algorithm could enumerate all the possible partitions in theory to obtain the best solution. But a heuristic method developed by Lloyd \cite{lloyd} offers a simple solution. Lloyd’s algorithm starts with K arbitrary centers, typically chosen uniformly at random from the data points \cite{lloyd1}. Each point is then assigned to the nearest center, and each center is recalculated as the mean of all points assigned to it. These two steps are repeated until a predefined termination criterion is met. The classical k-means algorithm is given in Algorithm 1 (\textbf{bold} symbols denote vectors). Here, $m[i]$ denotes the membership of point $\textbf{x}_i$, i.e. index of the cluster center that is nearest to $\textbf{x}_i$.

\begin{algorithm2e}
	\setstretch{1.35}
	\SetKwInOut{Input}{Input}
	\SetKwInOut{Output}{Output}
	
	\Input{$X = \{\textbf{x}_{1}, \textbf{x}_{2},\dots, \textbf{x}_{N}\} \in {\rm I\!R}^D$ ($ N \times D$ input data set)}
	\Output{$C = \{\textbf{c}_{1}, \textbf{c}_{2},\dots, \textbf{c}_{K}\} \in {\rm I\!R}^D$ (K cluster centers)}
	Select a random subset C of X as the initial set of cluster centers\;
	\While{termination criterion is not met}{
		\For{$(i = 1; i \le N; i = i+1)$}{
			Assign $x_i$ to the nearest cluster\;	
			$m[i] = \argmin_{k \in \{1,2,\dots,K\}} \begin{Vmatrix} \textbf{x}_i - \textbf{c}_k\end{Vmatrix}^2;$
		}
		Recalculate the cluster centers\;
		\For{$(k = 1; k \le K; k = k+1)$}{
			Cluster $S_k$ contains the set of points $\textbf{x}_i$ that are nearest to the center $\textbf{c}_k$\;	
			$S_k = \{\textbf{x}_i|m[i] = k\};$
			
			Calculate the new center $\textbf{c}_k$ as the mean of the points that belong to $S_k$;
			$\textbf{c}_k =\frac{1}{|S_k|}  \displaystyle\sum_{\textbf{x}_i \in S_k} \textbf{x}_i;$
		}	
	}
	\caption{Conventional K-Means Algorithm}
\end{algorithm2e}

\subsection{Proposed Image Compression Algorithm using k-Means}
\begin{figure*}
	\includegraphics[width=0.75\textwidth]{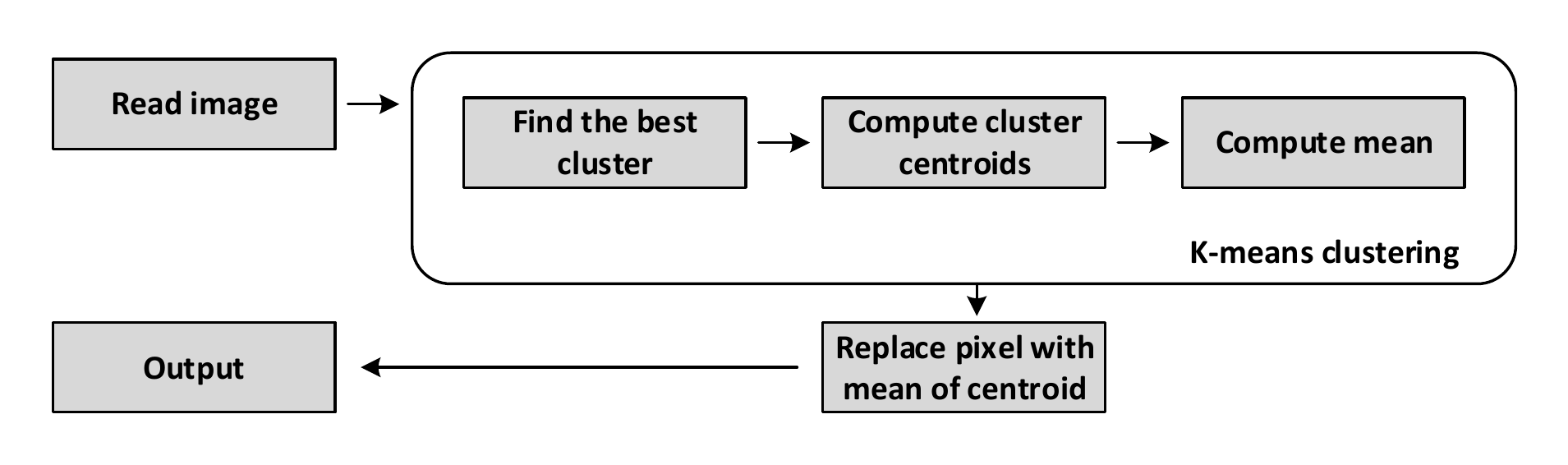}
	\centering
	\caption{Image compression module, following the procedure given in Algorithm 2.}
	\label{fig_image_compression_architecture}       
\end{figure*}

\begin{algorithm2e}
	\setstretch{1.35}
	\SetKwInOut{Input}{Input}
	\SetKwInOut{Output}{Output}
	
	\Input{Real image, $\mu$ number }
	\Output{Reconstructed image}
	
	\hspace{1cm}1. Load the color image $A(m \times n)$; \\
	\hspace{1cm}2. Read every pixel in the original image as sample data;\\
	\hspace{1cm}3. Apply k-Means algorithm to find the $\mu$ colors (clusters) that best group the pixels in the 3-dimensional RGB space;\\
	\hspace{1cm}4. Compute the cluster centroids again;\\
	\hspace{1cm}5. Assign each pixel position to its closest centroid;\\
	\hspace{1cm}6. Compute the mean of all the centroid;\\
	\hspace{1cm}7. Replace each pixel location with the mean of the assigned centroid;\\
	
	\caption{Image Compression Algorithm using k-Means Clustering}
\end{algorithm2e}

2D images typically use either 8-bit or 24-bit color. When using 8-bit color, there is a definition of up to 256 colors forming a palette for this image, each color denoted by an 8-bit value. A 24-bit color scheme, as the term suggests, uses 24 bits per pixel and provides a much better set of colors. For our purposes we'll consider 24 bit RGB (R'G'B') to be uncompressed color. The number 24 refers to the total number of bpp (bits per pixel) used to describe color. A 24 bit number has a range of 0 (no color or black) to 16,777,215 for a total of 16,777,216 different colors. Since Red, Green, and Blue are each described by 8 bits (24/3), which gives a range of 0 (no Red, no Green, or no Blue) to 255 for a total of 256 possible variations of each primary color. From this we can start to get an idea that the real-image contains thousands of colors and reduce the number of colors as a suitable number of colors. It can be 8, 16 or 32 colors, depend on the user demand and the network condition which is present in the next section. Procedure of image compression algorithm is shown in Figure~\ref{fig_image_compression_architecture}.

\begin{figure*}
	\includegraphics[width=1\textwidth]{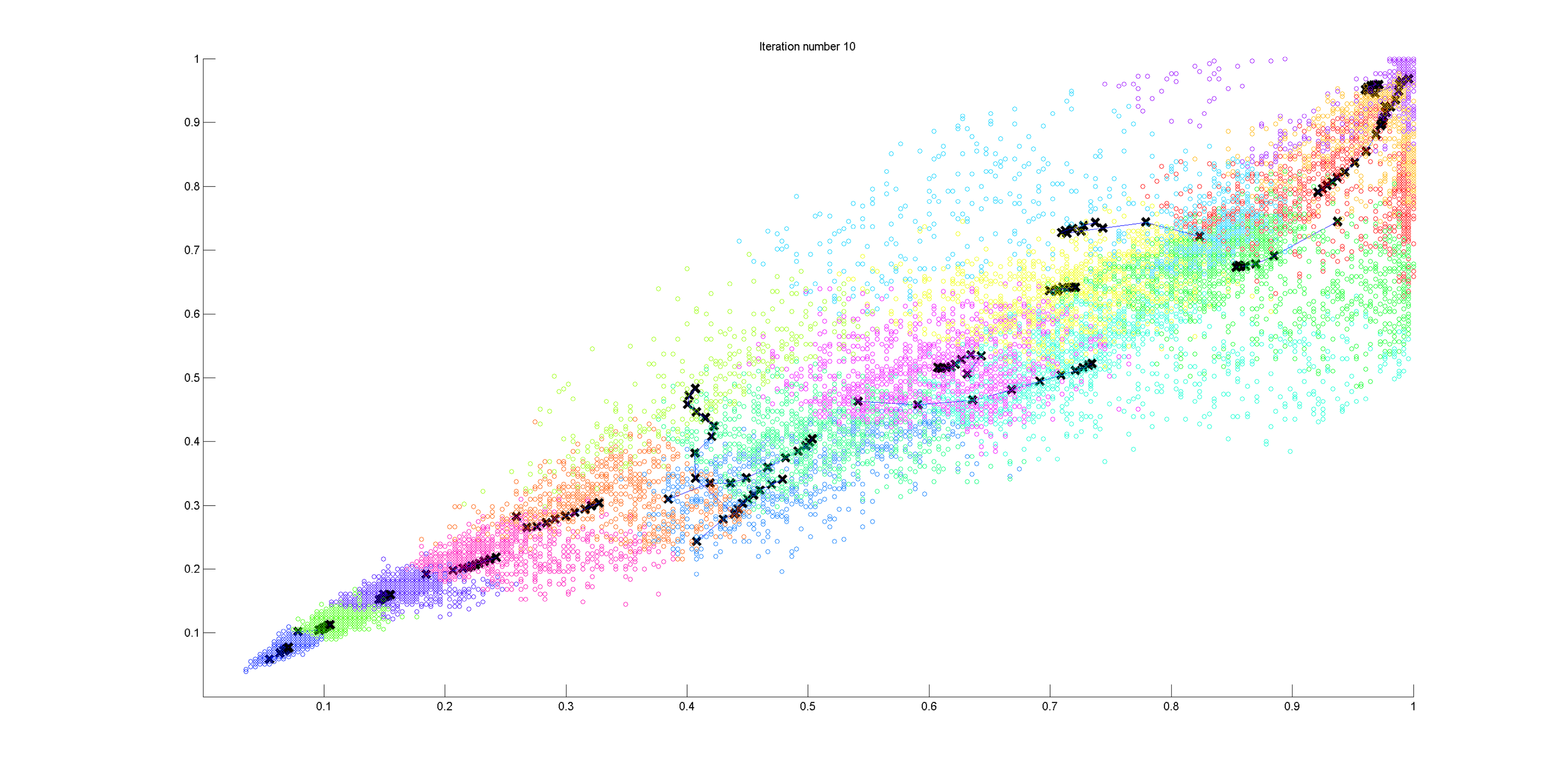}
	\centering
	\caption{Applying K-means clustering with a 2D image: the centroids stopped moving at the 10th iteration, meeting the convergence termination criterion. 16 colors (cluster) that best group be created}
	\label{fig_cluster}       
\end{figure*}

By making this reduction, it is possible to represent (compress) the photo in an efficient way. For example, suppose we want to reduce the number of colors to 16 of $256 \times 256$ image . Specifically, we need to store the RGB values of the 16 selected colors, and for each pixel in the images we now need to only store the index of the color at that location (where only 4 bits are necessary to represent 16 possibilities). Concretely, we will treat every pixel in the original image as a data example and use the k-means algorithm to find the 16 colors that best group (cluster) the pixels in the 3 dimensional RGB space. Figure~\ref{fig_cluster} shows that there are 16 groups of colors be created after applying the K-means clustering algorithm. First we have computed the cluster centroids on the image and then assigned each pixel position to its closest centroid. After that, we can view the effects of the compression by reconstructing the image based on the centroid assignments. Specifically, replace each pixel location with the mean of centroid assigned. Eventually, notice that after these steps we have significantly reduced the number of bits that are required to describe the image. The results can be summarized as follows: The original image required 24 bits for each one of the  $256 \times 256$ pixel locations, so total size of $256 \times 256 \times 24 = 1,572,864$ bits. The new representation requires some overhead storage in form of a dictionary of 16 colors, each of which require 24 bits, but the image itself then only requires 4 bits per pixel location. The final number of bits used is therefore $16 \times 24 + 256 \times 256 \times 4 = 262,528$ bits, which corresponds to compressing the original image by about a factor of 6.
  
\section{System Architecture}

\begin{figure*}
	\includegraphics[width=1\textwidth]{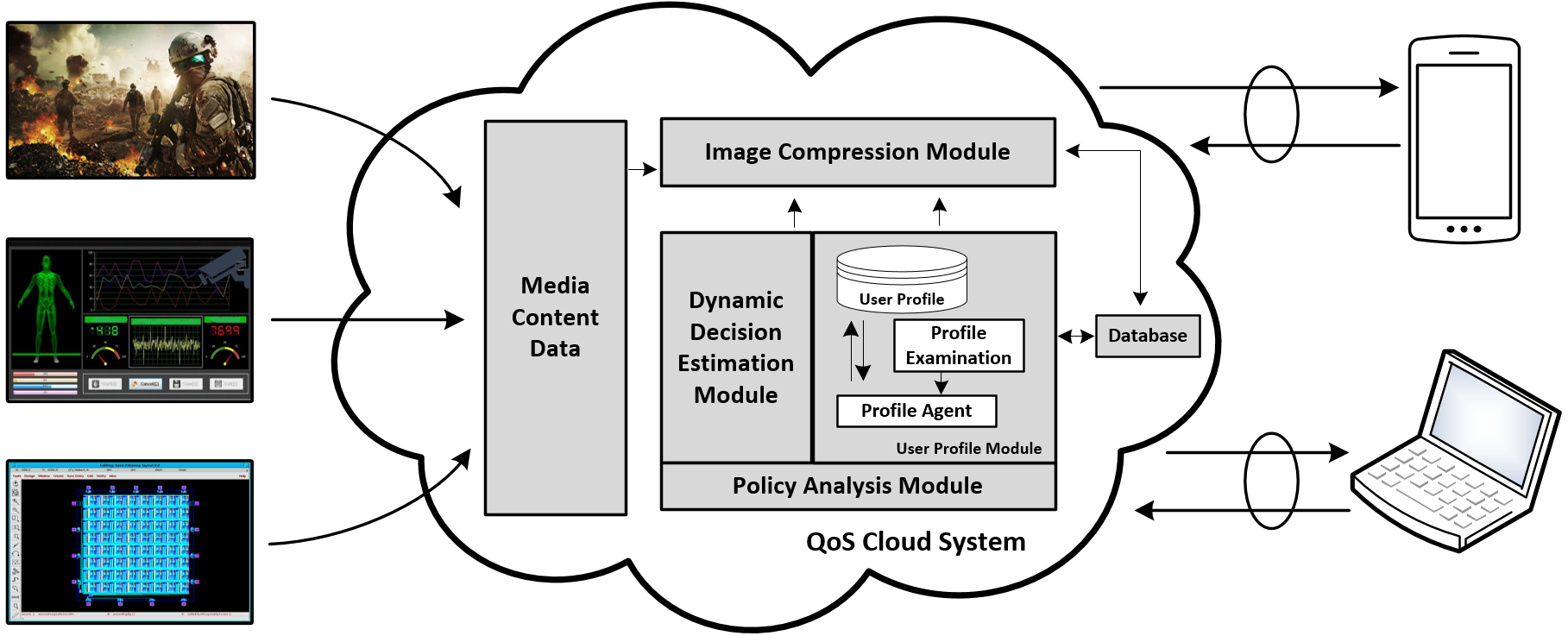}
	\centering
	\vspace{-0em}
	\caption{Proposed system architecture of Cloud-based remote desktop streaming}
	\label{fig_system_architecture}       
\end{figure*}

The proposed system provides an efficient interactive streaming desktop remote service for diversified client devices and dynamic network environments. The overall system structure is shown in Figure~\ref{fig_system_architecture}. Whenever a device requests a multimedia desktop streaming service, by the agent is installed on the client device, it helps transmit data included hardware and network environment parameters to User Profile Module (UPM). After that the UPM records the device codes and determines the required parameters, and then transmits them to the Dynamic Decision Estimated Module (DDEM). By analysis of historical data, the DDEM determines the most suitable compression ratio $\mu$ number for the device according to the parameters and the network condition. The $\mu$ number is transferred to the Image Compression Module (ICM) for use in the encode phase. Finally, the multimedia video file is transmitted to the mobile device through the service.

\subsection{User Profile Module}

\begin{table}[!h]
	\centering
	\caption{Data Structure of Feature Parameters}
	\label{data_structure_parameter}       
	\begin{tabular}{lll}
		\hline\noalign{\smallskip}
		Data Type & Function & Description  \\
		\noalign{\smallskip}\hline\noalign{\smallskip}
		int & Resolution & Supportable resolution \\
		float & CPU & Operation ratio \\
		float & Battery & Dump energy of the device \\
		float & Bandwidth & Tested existing bandwidth \\
		\noalign{\smallskip}\hline
	\end{tabular}
\end{table}

The profile agent is used to receive the device hardware environment parameters and create a user profile. The device transmits its hardware specifications in XML-schema format to the profile agent in the cloud server, as shown in the Table~\ref{data_structure_parameter}. The XML-schema is meta-data, which is mainly semantic and assists in describing the data format of the file. The meta-data enables non-owner users to see information about the files, and its structure is extensible. However, any device that is using this cloud service for the first time will be unable to provide such a profile, so there shall be an additional profile examination to provide the test performance of the mobile device and sample relevant information. Through this function, the client device can generate an XML-schema profile and transmit it to the profile agent. The profile agent determines the required parameters for the XML-schema and creates a user profile, and then transmits the profile to the DDEM for identification.

\subsection{Dynamic Decision Estimation Module with QoS Control}

\begin{figure*}
	\includegraphics[width=0.85\textwidth]{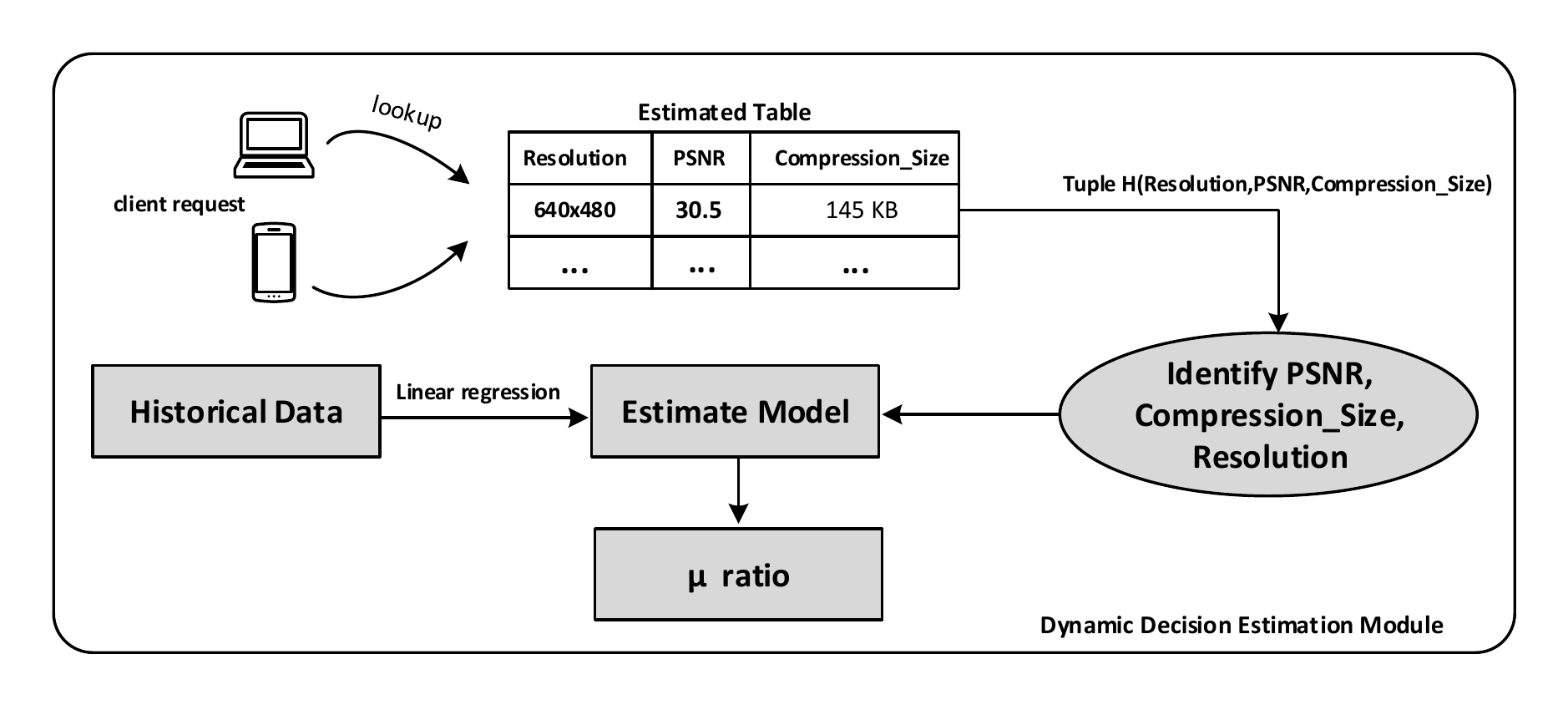}
	\centering
	\caption{Model of Dynamic Decision Estimation Module}
	\label{fig_ddem}       
\end{figure*}

Figure~\ref{fig_ddem} illustrates the DDEM used to estimate the value of $\sigma$. Specifically, the DDEM conducts this estimation after receiving the profile data from the User Profile Module. Based on multiple linear regression model in statistics, Based on multiple linear regression model in statistics, the DDEM determines the relationship between factors and predicts a suitable range for the number $\mu$ number. The problem is how to choose the best $\mu$ to obtain a suitable image size after the compression phase while also ensuring sufficiently high image quality and satisfying the image resolution demanded by the user. The ICM is used to obtain the compression data and calculate the corresponding $PSNR$ values; these data are defined as historical data for the simulations, as shown in Table~\ref{table_compression_data}. 

In the current application, PSNRs of 25 to 50 dB are considered to be acceptable in terms of transmission quality. Therefore, before the regression analysis, we need to extract a list of values filtered by the condition $PSNR \ge 25$. After we have a suitable range for the $\mu$ ratio, the system will analyze which $\mu$ number is most suitable for the compression phase. Specifically, let $\sigma$ be the available bandwidth and $\theta$ the bandwidth threshold. If $\sigma$ is greater than $\theta$, the system can transmit the video content with the best possible quality without further processing. Otherwise, the next step is to determine appropriate values of $PSNR$ and  $Compression\_Size$ to adapt to the current network conditions. In this case, the $PSNR$ and $Compression\_Size$ values are looked up in the estimation table; this table comprises a set of tuples of the form $h(Resolution, PSNR, Compression\_Size)$, as described in the next section. Eventually, these tuples obtained is input to the estimate model, then the output $\mu$ value is asymptotic as well. The operation of QoS control is described algorithmically as Algorithm 3.
 
\begin{algorithm2e}
	\setstretch{1.35}
	\SetKwInOut{Input}{Input}
	\SetKwInOut{Output}{Output}
	
	\Input{$PSNR, Compression\_Size, Resolution, Network\_Condition$}
	\Output{$\mu$ compression ratio}
	Filter historical data with $25 \le PSNR \le 50$ \;
	In historical data, model the relationship between input factors by means of linear regression\;
	\While{streaming video}{
		\eIf{available bandwidth $\sigma$ is greater than a predefined threshold $\theta$}{
			Full transmission;
		}{
		Adjust $\sigma$ value based on consideration of the network bandwidth\;
		Identify the $PSNR$ and $Compression\_Size$ suitable for the given $\sigma$ while ensuring sufficiently high image quality\;
		Input the $PSNR, Compression\_Size, Resolution$ values to the linear model\;
		Get the best $\mu$ compression ratio and transfer it to the ICM;
	}	
}
\caption{QoS Control System Algorithm}
\end{algorithm2e}
\subsubsection{Proposed Estimation Model}
In the present paper, we use multiple linear regression (MLR) to estimate the value of a variable based on the value of two or more other variables. The goal of MLR is to model the statistical relationship between response and predictor variables. A simple linear regression illustrates the relation between the dependent variable $y$ and the independent variable $x$ based on the regression equation

\begin{equation}\label{Eq3}
y_i = \beta_0 + \beta_1x_i +  e_i, \quad i = 1,2,3,\dots, n
\end{equation}

According to the multiple linear regression model the dependent variable is related to two or more independent variables. The general model for k variables is of the form

\begin{equation}\label{Eq4}
y_i = \beta_0 + \beta_1x_{i1} + \beta_2x_{i2} + \dots + \beta_kx_{ik} + e_i,\quad i = 1,2,3,\dots, n
\end{equation}

To simplify the computation, the multiple regression model in terms of the observations can be written using matrix notation. Using matrices allows for a more compact framework in terms of vectors representing the observations, levels of regression variables, regression coefficients and random errors. The model is in the form

\begin{equation}\label{Eq5}
Y = X\beta + \epsilon
\end{equation}

and when written in matrix notation we have 

\begin{equation}\label{Eq6}
\begin{bmatrix} y_1 \\ y_2 \\ \vdots \\ y_n \end{bmatrix} = \begin{bmatrix} 1 & x_{11} & \cdots & x_{1k} \\ 1 & x_{21} & \cdots & x_{2k} \\ \vdots & \vdots & \ddots & \vdots \\ 1 & x_{n1} & \cdots & x_{nk}\end{bmatrix}\begin{bmatrix} \beta_1 \\ \beta_2 \\ \vdots \\ \beta_n \end{bmatrix} + \begin{bmatrix} \epsilon_1 \\ \epsilon_2 \\ \vdots \\ \epsilon_n \end{bmatrix}
\end{equation}

where $Y$, $X$ denote respectively the response values for all observations into a $n$-dimensional vector and predictor into a $n \times p + 1$ matrix. $\beta$ denote the intercepts and slopes into a $k+1$-dimensional vector and $\epsilon$ is an $n \times 1$ vector of random errors.

The first step in multiple linear regression analysis is to determine the vector of least squares estimators $\widehat{\beta}$ which gives the linear combination $\widehat{y}$ that minimizes the length of
the error vector. Basically the estimator $\widehat{\beta}$ provides the least possible value to sum of the squares difference between $\widehat{y}$ and $y$. Now, since the objective of multiple regression is to minimize the sum of the squared errors, the regression coefficients that meet this condition are determined by solving the least squares normal equation

\begin{equation}\label{Eq7}
X^TX\widehat{\beta} = X^TY
\end{equation}

Now if the variables $x1, x2,\dots, xn$ are linearly independent, then the inverse of $X^TX$, namely $(X^TX)^-1$ will exist. Multiplying both sides of the normal Eq.~\ref{Eq7} by $(X^TX)^-1$, we obtain 

\begin{equation}\label{Eq8}
\widehat{\beta} = (X^TX)^{-1}X^TY
\end{equation}

In the present work, $\mu$ is considered to be the response variable and the \textit{$Compression\_Size$}, \textit{$Resolution$} and \textit{$PSNR$} are used as predictor variables. In this case, we defined the $Resolution$ as a value from 1 to 7 corresponding to one of seven common screen resolutions; these resolutions are listed in Table~\ref{table_compression_data}. Thus, from Eq.~\ref{Eq4}, we have the following linear model for this approach.

\begin{equation}\label{Eq9}
\mu = \beta_0 + \beta_1\ast\textit{$Compression\_Size$} + \beta_2\ast\textit{$Resolution$} + \beta_3\ast\textit{$PSNR$}
\end{equation}
By filtering the historical data using the condition $25 \le PSNR \le 50$, we can now find the least squares estimators $\widehat{\beta}$ by using the Eq.~\ref{Eq8}. Finally, the resulting regression model is as follows.

\begin{equation}\label{Eq10}
	\begin{array}{l}
		\widehat{\mu} = \widehat{\beta_0} + \widehat{\beta_1}\ast\textit{$Compression\_Size$} + \widehat{\beta_2}\ast\textit{$Resolution$} + \widehat{\beta_3}\ast\textit{$PSNR$}\\
		= -310.72 + 0.051\ast\textit{$Compression\_Size$} \\+  \quad 5.7905\ast\textit{$Resolution$} + 11.158 \ast \textit{$PSNR$}
	\end{array}
\end{equation}

\begin{figure*}
	\includegraphics[width=0.7\textwidth]{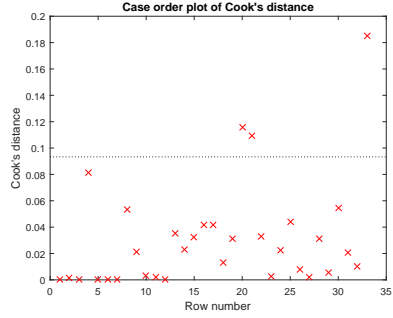}
	\centering
	\caption{Data points with large Cook's distances}
	\label{cook_distance}       
\end{figure*}

\begin{figure}[h!]
	\begin{center}
		$\begin{array}{cc}
		\includegraphics[width=2.4in]{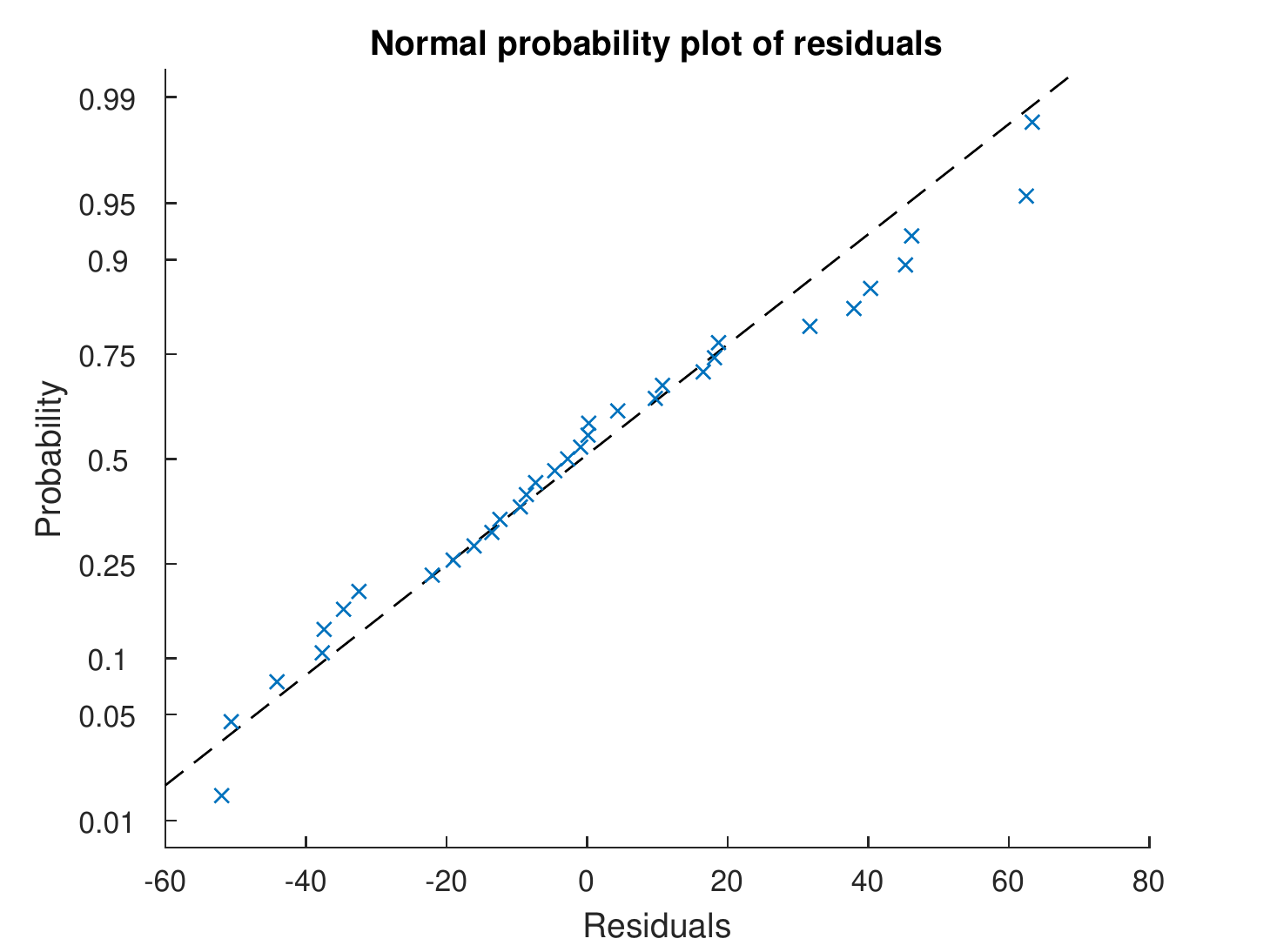} & \includegraphics[width=2.4in]{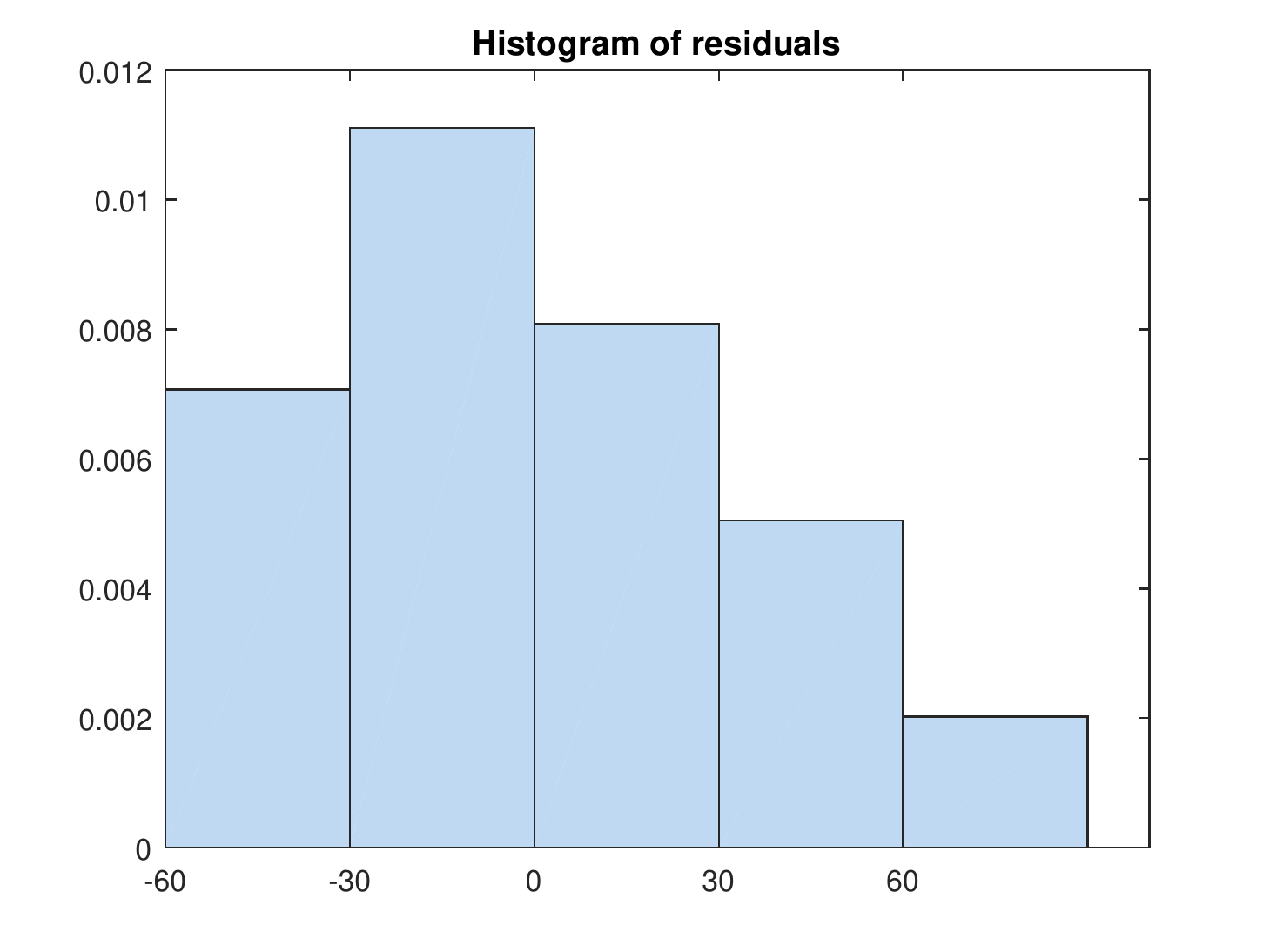}\\
		(a) & (b)
		\end{array}$
	\end{center}
	\caption{Histogram and Normal Probability Plot of Residuals after removing outliers}
	\label{his_pro}
\end{figure}

In linear regression, one problem with least squares occurs when there are one or more large deviations, i.e. cases whose values differ substantially from the other observations. These points are called outliers. There are various methods for outlier detection \cite{outlier}. As shown in Figure~\ref{cook_distance}, using Cook's distance which is commonly used of the influence of a data point when performing least squares regression analysis. Cook's distance used to indicate data points that are particularly worth checking for validity. In this case, there are several outliers above the reference line; these are identified and removed from the model. 

After reducing the outliers, we obtain the following fitted regression model that represents the data adequately.

\begin{equation}\label{Eq11}
\begin{array}{l}
\widehat{\mu} = -313.97 + 0.0496\ast\textit{$Compression\_Size$} \\+  \quad 4.1217\ast\textit{$Resolution$} + 11.444 \ast \textit{$PSNR$}
\end{array}
\end{equation}

The difference between the observed and predicted score, Y-Y', is called a residual. We analyze a histogram that shows the range of the residuals and their frequencies, and a probability plot that shows how the distribution of the residuals compares to a normal distribution with matching variance. As shown in Figure~\ref{his_pro}, the new residuals plot looks fairly symmetric and the probability plot seems reasonably straight, suggesting that the data fits reasonably to normally distributed residuals.

\subsubsection{Estimated Table}
At this point, we have obtained an expected range of $\mu$ values from 8 to 64 while ensuring that $25 \le PSNR \le 50$. Therefore, to estimate a suitable $\mu$ value, we need to define an estimation table comprising a set of tuples of the form $h$($Resolution$, $PSNR$, $Estimated\_Size$), where the $Estimated\_Size$ value is selected according to the average compressed image size of each group of images ($8 \le \mu \le 64$) in Table~\ref{table_compression_data}. Moreover, by analyzing the historical data, we have classified the $Resolution$ further as mobile thin resolutions and desktop resolutions. $Resolution = 1$ denote the thin client resolutions, $Resolution = 2$ denote the desktop resolutions. These are listed in Table~\ref{my-label}.

\begin{table}[!h]
	\centering
	\caption{Estimation Table}
	\label{my-label}
	\begin{tabular}{|l|l|l|l|}
		\hline
		& Resolution & PSNR & Estimated\_Size\ \\ \hline
		\multirow{3}{*}{Thin Client} &      300 x 212      &  28    &        20.6 kB          \\ \cline{2-4} 
		&         $\dots$      &  $\dots$     &       $\dots$              \\ \cline{2-4} 
		&    600 x 450        &    30  &       42.2 kB            \\ \hline
		\multirow{3}{*}{Desktop}     &       800 x 600     &   25   &            373.9 kB       \\ \cline{2-4} 
		&       $\dots$     &  $\dots$     &         $\dots$           \\ \cline{2-4} 
		&      1920 x 1080      &   40   &        657.2 kB           \\ \hline
	\end{tabular}
\end{table}
%
%
Specifically, the system will automatically adjust to increase or decrease the PSNR according to the available bandwidth $\sigma$ representing the current network status. Then, by looking up the predictor variables $PSNR$, $Resolution$ and $Estimated\_Size$ in the estimated table, we can find the best tuple to input into the estimate model. Obviously, the $\mu$ value obtained here satisfies the client demands.

For example, consider a case in which a thin client device sends a request to the system, and the available bandwidth $\sigma$ is less than or equal to a predefined threshold value $\theta$ representing $80$ ò the full bandwidth. The system will then decrease the PSNR value. The default is $30$ dB. Assume we would decrease the PSNR down to $28$ dB. From the User Profile Module, the system will know what type of device the request is originating from. In this case, we get the tuple $h(28, 1, 20.6)$ from the table. Thus, we obtain the following.

\begin{equation}\label{Eq12}
\begin{array}{l}
\widehat{\mu} = -313.97 + 0.0496 \ast 20.6 +  4.1217\ast 1 + 11.444 \ast 28  = 11.60546
\end{array}
\end{equation}

Then resulting $\mu = 11.60546$ is exactly the expected compression ratio; this value is input to the ICM. Thus, after the compression phase the output frame ensures the image quality as well as the screen resolution.

\section{Experimental Results}
\subsection{Image Compression Results}
The proposed method was tested on some images in a dataset collected by the MIT Computer Science and Artificial Intelligence Lab. The natural images in the set include Zebra Butterfly ($300\times212$, $27.4$ kB), Boston Street ($590 \times 430$, $069.2$ kB), Conference Room ($600\times450$, $61.4$ kB), Airport ($745\times258$, $46.2$ kB), Oxford Outdoor ($1526\times2048$, $1.21$ MB), Chip Designer ($1680\times1050$, $503$ kB), and Angry Bird ($1920\times1080$, $153$ kB). All of the image compression programs were implemented in Matlab 2010 and executed on an Intel Core i5-4670 CPU running at 3.40 GHz.

\begin{figure}[!h]
	\captionsetup[subfigure]{labelformat=empty}
	\centering
	\footnotesize
	
	\stackunder[5pt]{\includegraphics[width=2.85cm,height=2.85cm]{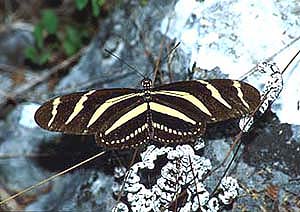}}{Butterfly}
	\stackunder[5pt]{\includegraphics[width=2.85cm,height=2.85cm]{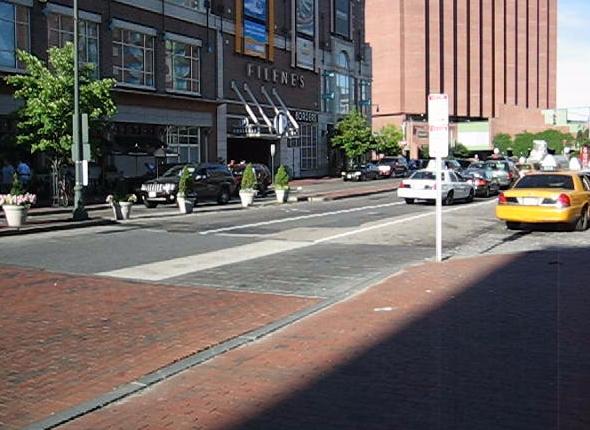}}{Boston Street}
	\stackunder[5pt]{\includegraphics[width=2.85cm,height=2.85cm]{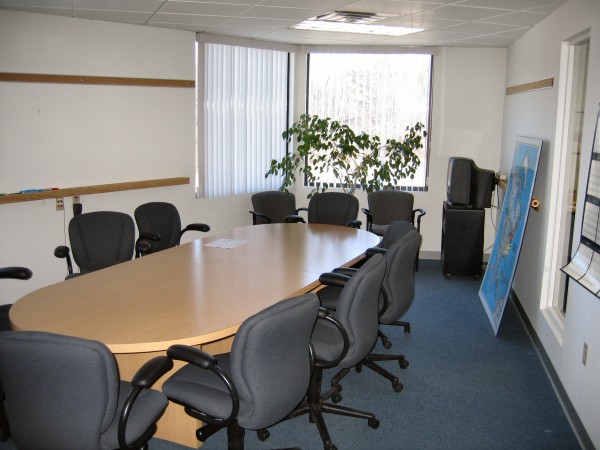}}{Room}
	
	\stackunder[5pt]{\includegraphics[width=2.85cm,height=2.85cm]{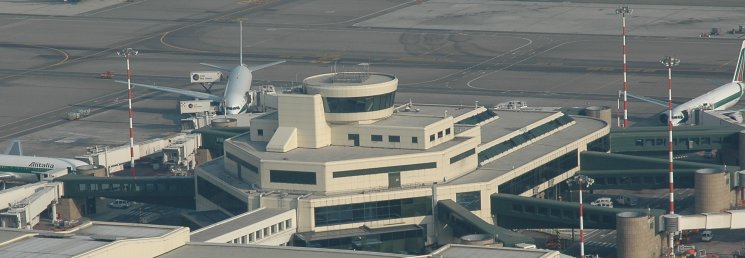}}{Airport}
	\stackunder[5pt]{\includegraphics[width=2.85cm,height=2.85cm]{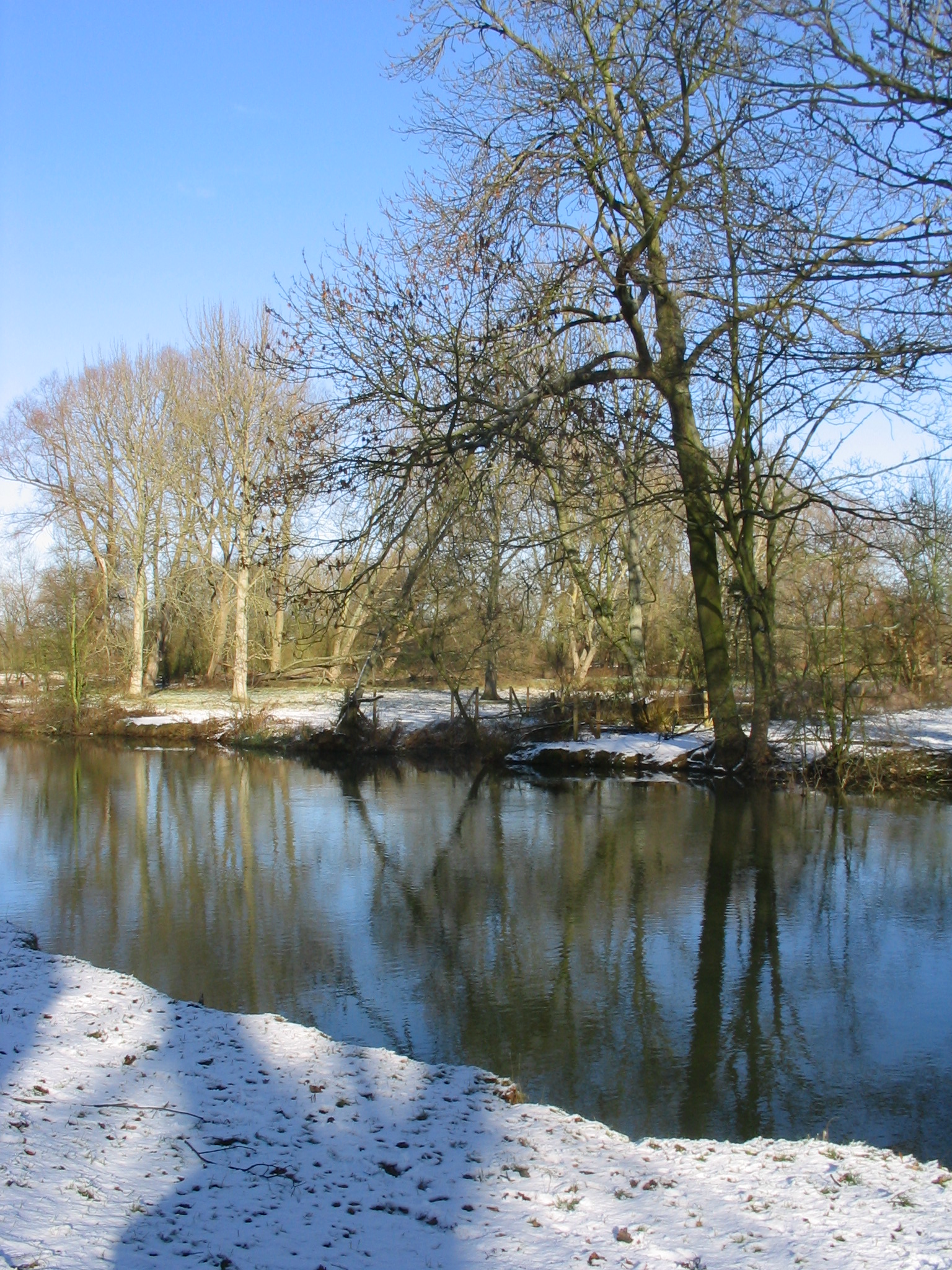}}{Oxford}
	\stackunder[5pt]{\includegraphics[width=2.85cm,height=2.85cm]{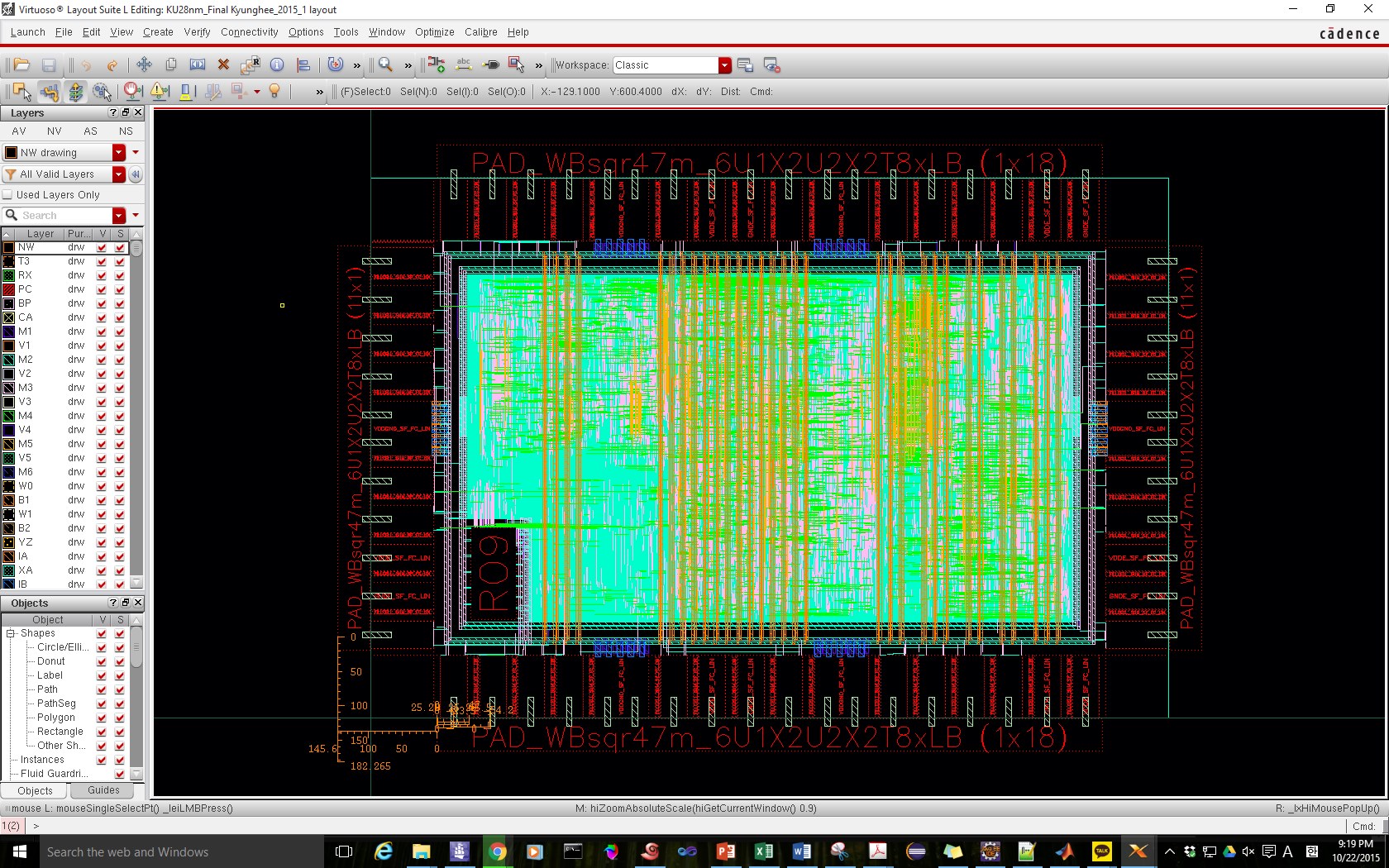}}{Chip Designer}
	\stackunder[5pt]{\includegraphics[width=2.85cm,height=2.85cm]{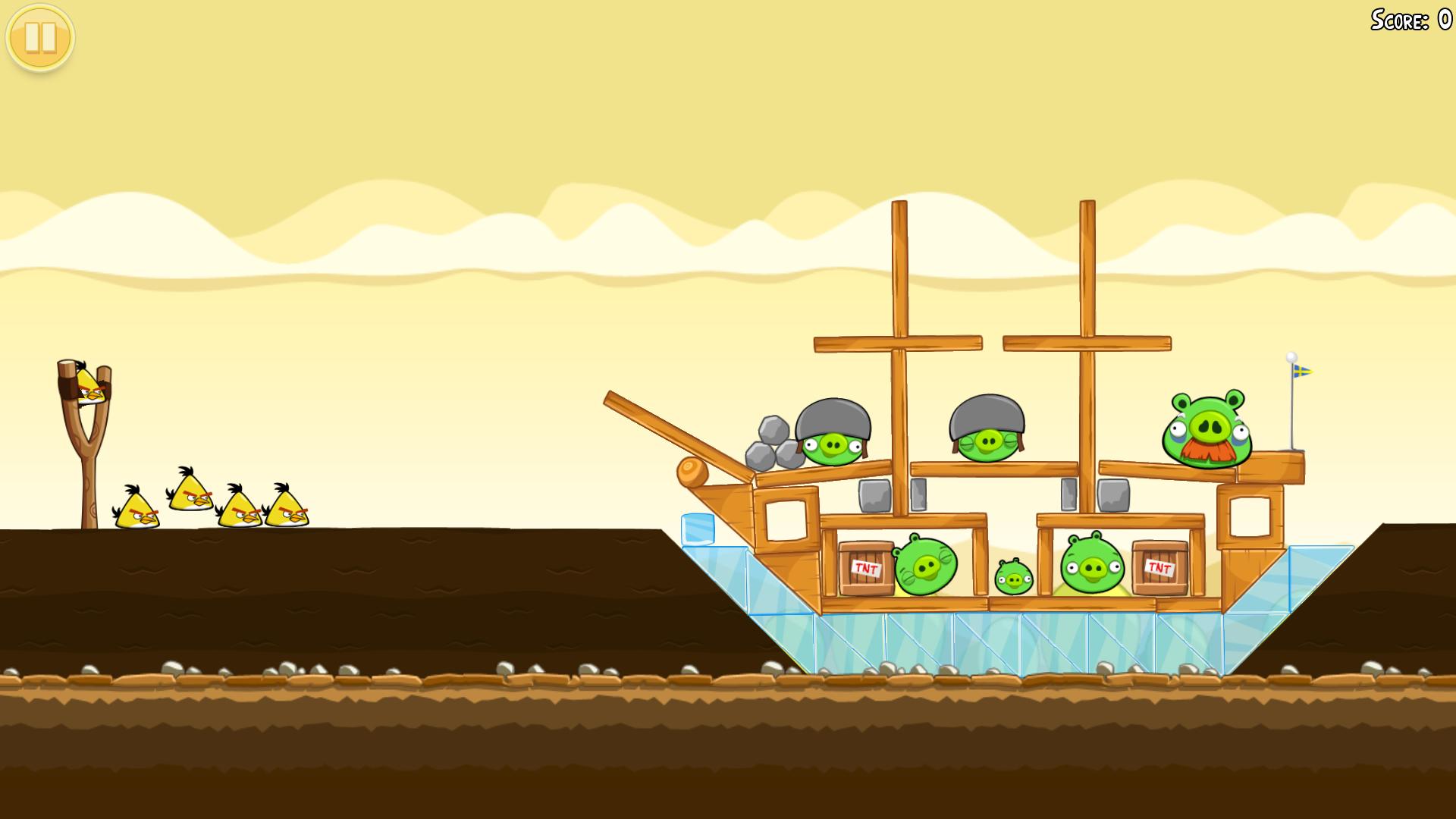}}{Angry Bird}
	
	\caption{Test images}
	\label{fig:input_image}
\end{figure}

The effectiveness of the compression was quantified by the mean squared error (MSE) measure:

\begin{equation}\label{Eq13}
MSE (X,\widehat{X}) = \frac{1}{HW}\displaystyle\sum_{h = 1}^{H}\displaystyle\sum_{w = 1}^{W} \begin{Vmatrix} x(h,w) - \widehat{x}(h,w)\end{Vmatrix}^2_2
\end{equation}

where $X$ and $\widehat{X}$ respectively denote the $H \times W$ of the original and compressed images in the RGB color space. MSE represent the average distortion with respect to the $L^2_2$ norm \cite{lloyd}. Note that the Peak Signal-to-Noise Ratio (PSNR) measure can be easily calculated from the MSE value:

\begin{equation}\label{Eq14}
PSNR = 20\log_{10}(\frac{255}{MSE})
\end{equation}

\begin{figure}[!h]
	\captionsetup[subfigure]{labelformat=empty}
	\centering
	\footnotesize
	
	\stackunder[5pt]{\includegraphics[width=5cm,height=4.5cm]{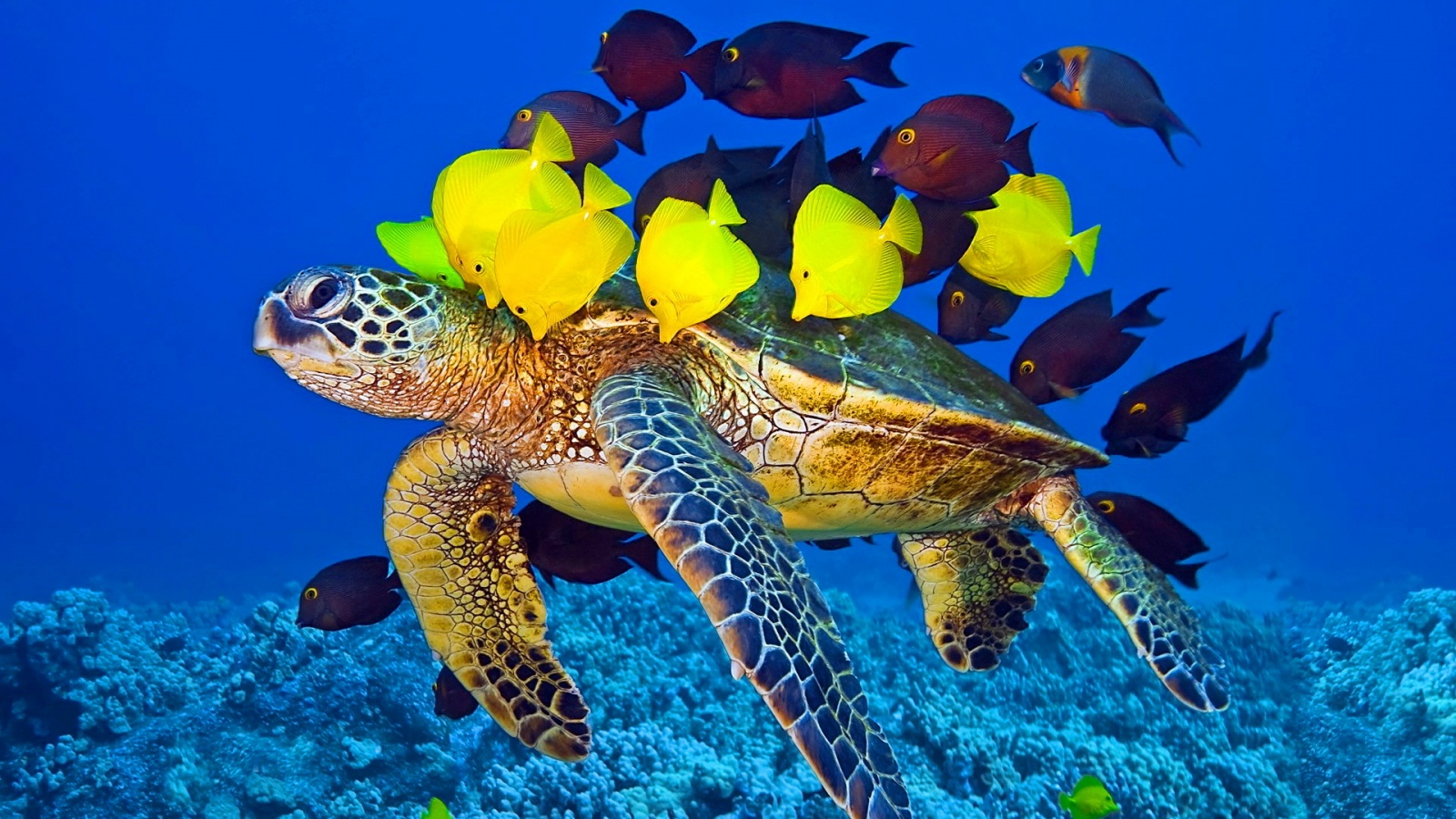}}{a) turtle.jpg - $514$ kB }
	\stackunder[5pt]{\includegraphics[width=5cm,height=4.5cm]{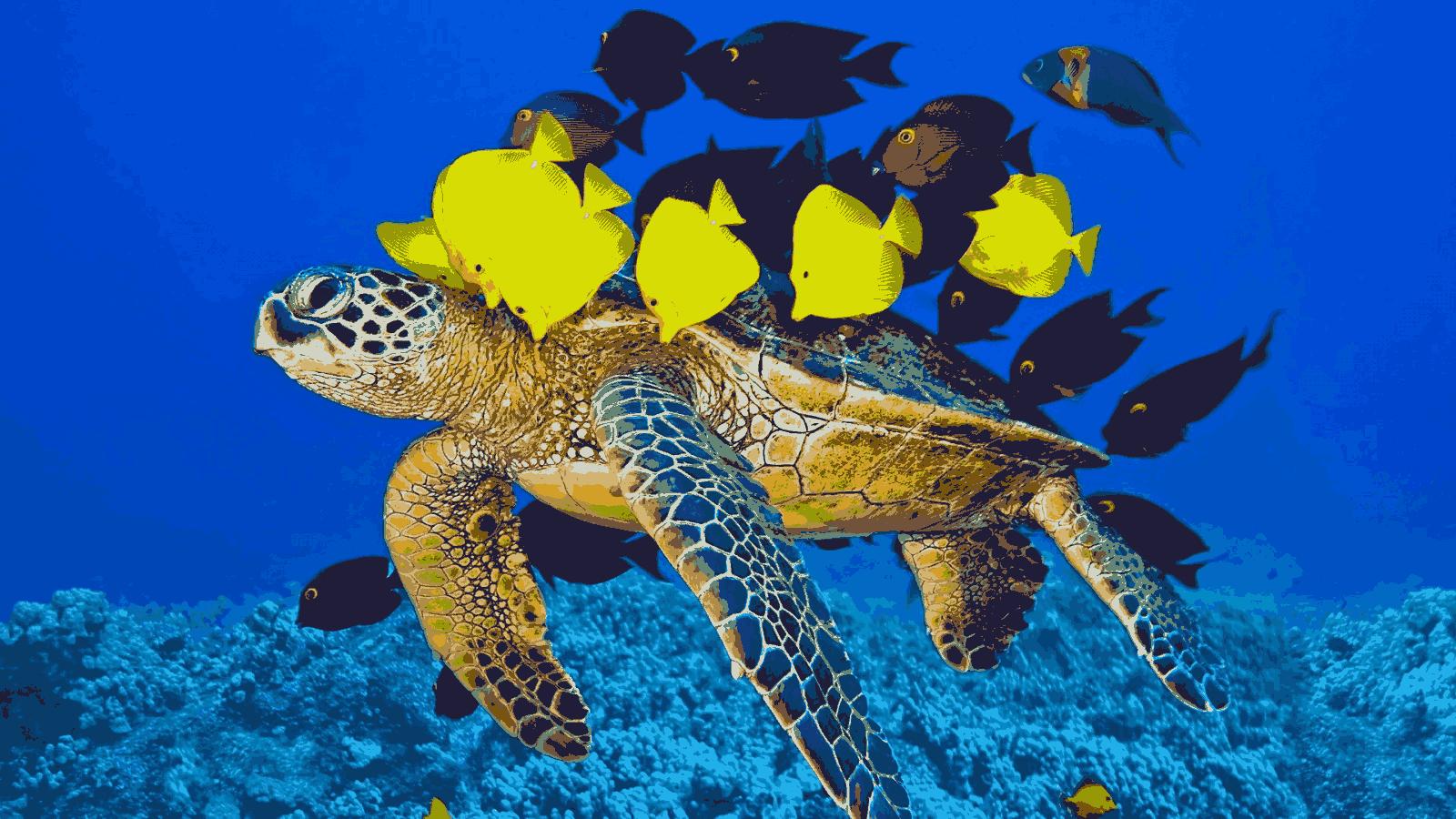}}{b) turtle\_32.jpg - $222$ kB 
		}
	
	\caption{Comparison of original image and compressed image with $\mu = 32$. The PSNR of compressed image is $26.5845$ dB.   }
	\label{fig:jpeg_comparison}
\end{figure}

Let $\alpha$ denote the color depth, the image dimensions $A(m \times n)$, and let $\mu$ be defined as the number of colors (clusters) and $\gamma$ as the number of bits necessary to represent the index of the colors at the location in the image. We computed the compression ratio (CR) by dividing the size of the original file in bits by the size of the compressed file:
\begin{equation}\label{Eq2}
CR = \frac{|G|}{\mu \times \alpha + m \times n \times \gamma}
\end{equation}

where $|G| = \alpha \times m \times n$ denotes the size of the original image. Figure~\ref{fig:jpeg_comparison} shows the original image Turtle and a compressed version of Turtle with $\mu = 32$. The compression was carried out using various values of $\mu$; the results are shown in Table~\ref{table_compression_data} and Figure~\ref{fig:output_image} .

As can be seen, the presently proposed approach enabled compressing an image to $70\%$ of its original size while maintaining $95\%$ visual similarity to the original image. It is of extremely importance to reduce the frame loss rate when transmitting video data over the Internet. If the frame loss rate is small, it means that the video streaming quality is good.  

Basically, if a packet is lost during transmission, a common method for error concealment on the decoder side is to replace the data lost by the last correctly received frame data. In this case, let $\widetilde{p}_n^i$ denote the $i^{th}$ pixel of the $n^{th}$ reconstructed frame at the decoder, and $p^i_n$ denote the $i^{th}$ pixel of the $n^{th}$ original coded frame at the encoder. The total frame error at frame $n$ is defined by

\begin{equation}\label{Eq15}
e_n = \displaystyle\sum_{i = 1}^{M}(\widetilde{p}_n^i - p^i_n)
\end{equation}

where M is the number of pixels in each frame (i.e $176x144=25344$ in QCIF video format). The Mean Square Error associated with frame error $e_n$ is

\begin{equation}\label{Eq16}
d_n = E[(\widetilde{p}_n^i - p^i_n)^2] = \frac{1}{M}\displaystyle\sum_{i = 1}^{M}[(\widetilde{p}_n^i - p^i_n)^2]
\end{equation}
\begin{figure*}
	\includegraphics[width=0.85\textwidth]{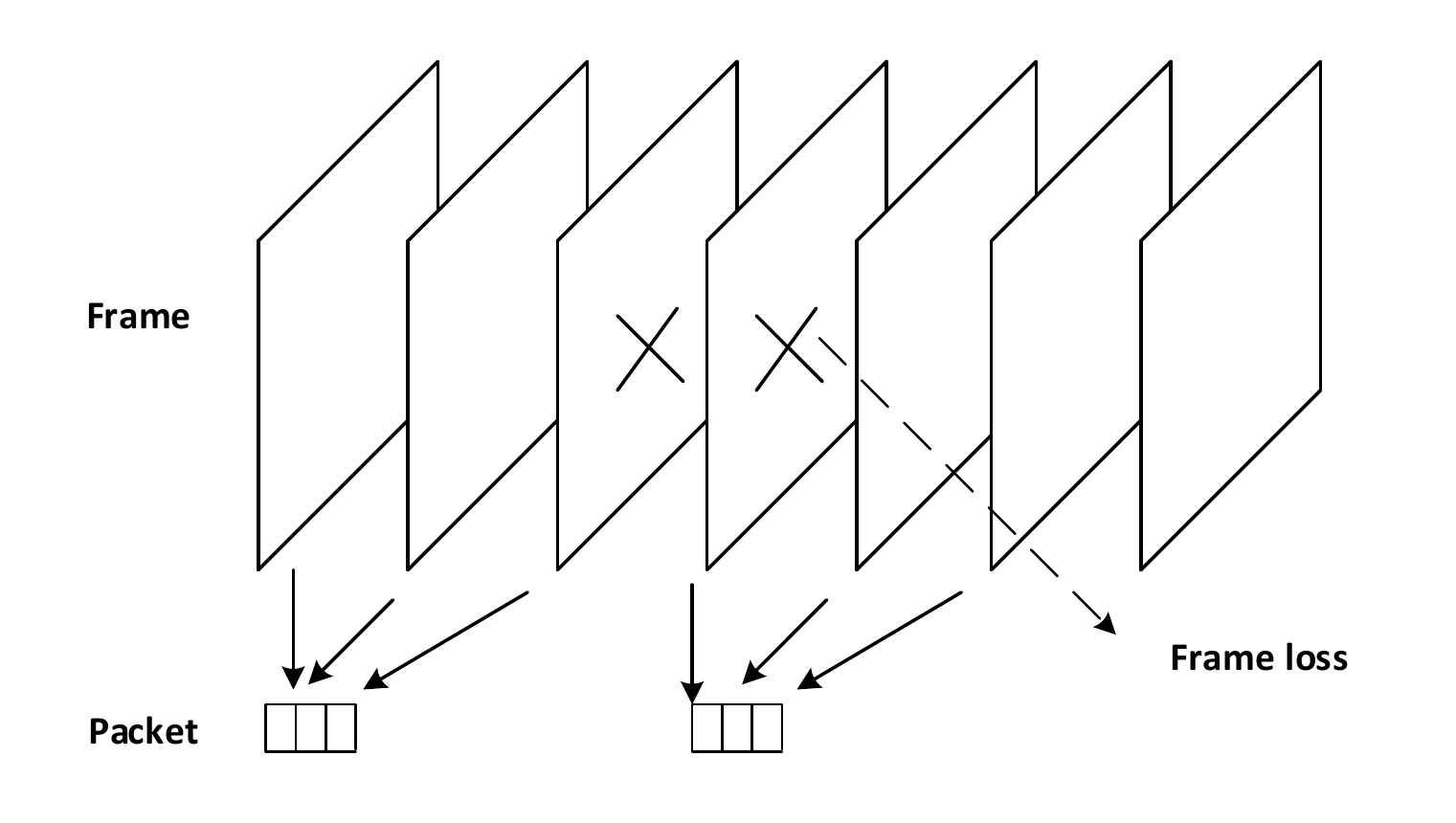}
	\centering
	\caption{Model of Dynamic Decision Estimation Module}
	\label{fig_frameloss}       
\end{figure*}

Then, the PSNR of the video signal of frame $n$ is given by 

\begin{equation}\label{Eq17}
PSNR_{dB}[n] = 20log_{10}(\frac{V_{peak}}{RMSE}) = 20log_{10}(\frac{V_{peak}}{\sqrt{d_n}})
\end{equation}

where $V_{peak}$ is the maximum possible pixel value of the frame and RMSE is the root mean square error between received and original frames. In practice, if the frame size is small then the distortion arising from the loss of a single frame is small too. From the form of Eq.~\ref{Eq17}, it is clear that $d_n$ decreases as $PSNR_n$ increases. So, it is a important problem we need to consider when ensuring performance of the compression.
\subsection{Evaluation of QoS System}
\begin{figure}[!h]
	\captionsetup[subfigure]{labelformat=empty}
	\centering
	\footnotesize
	
	\stackunder[5pt]{\includegraphics[width=5.5cm,height=4cm]{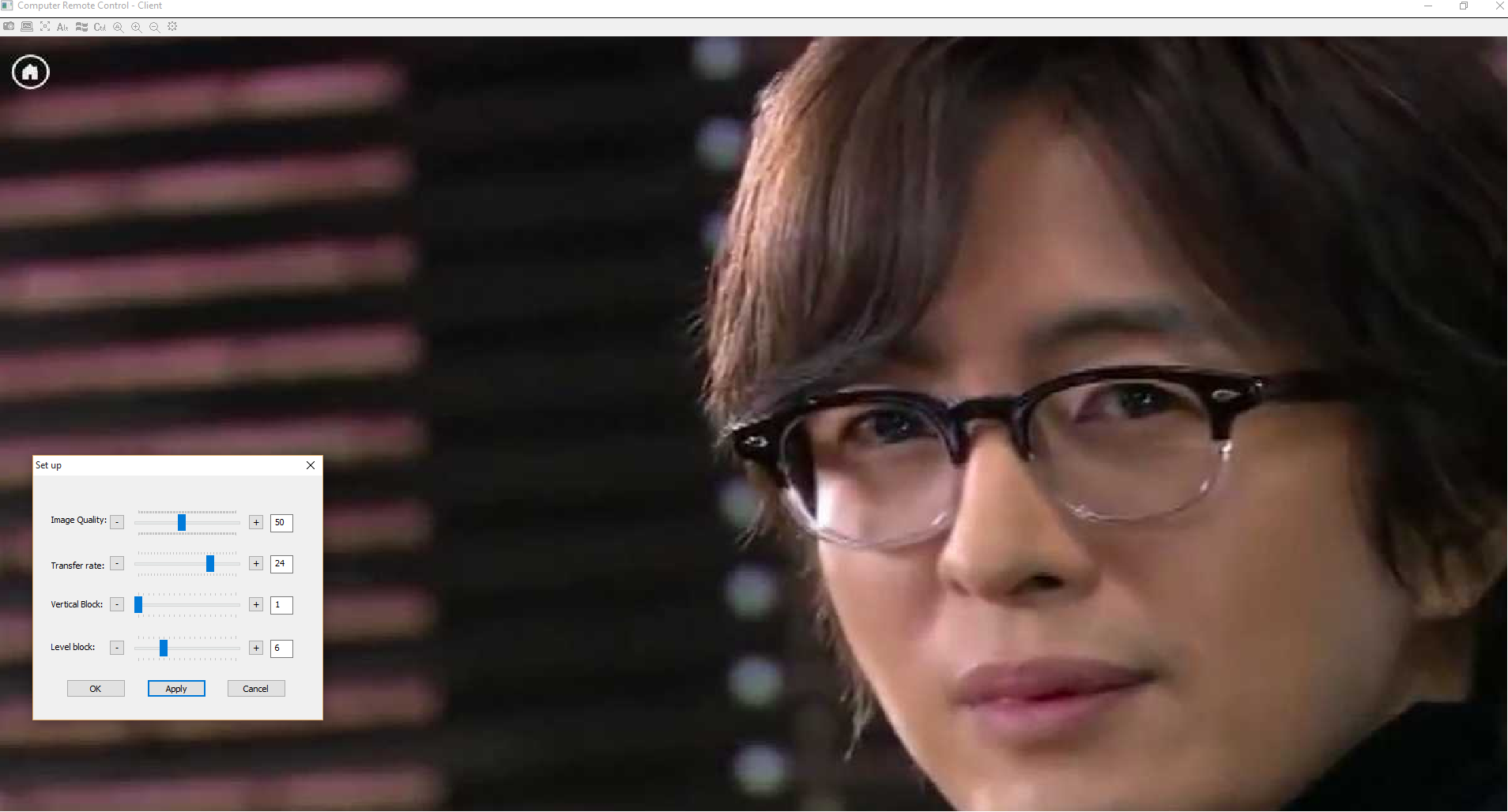}}{a) A Scene in the Korean Movie }
	\stackunder[5pt]{\includegraphics[width=5.5cm,height=4cm]{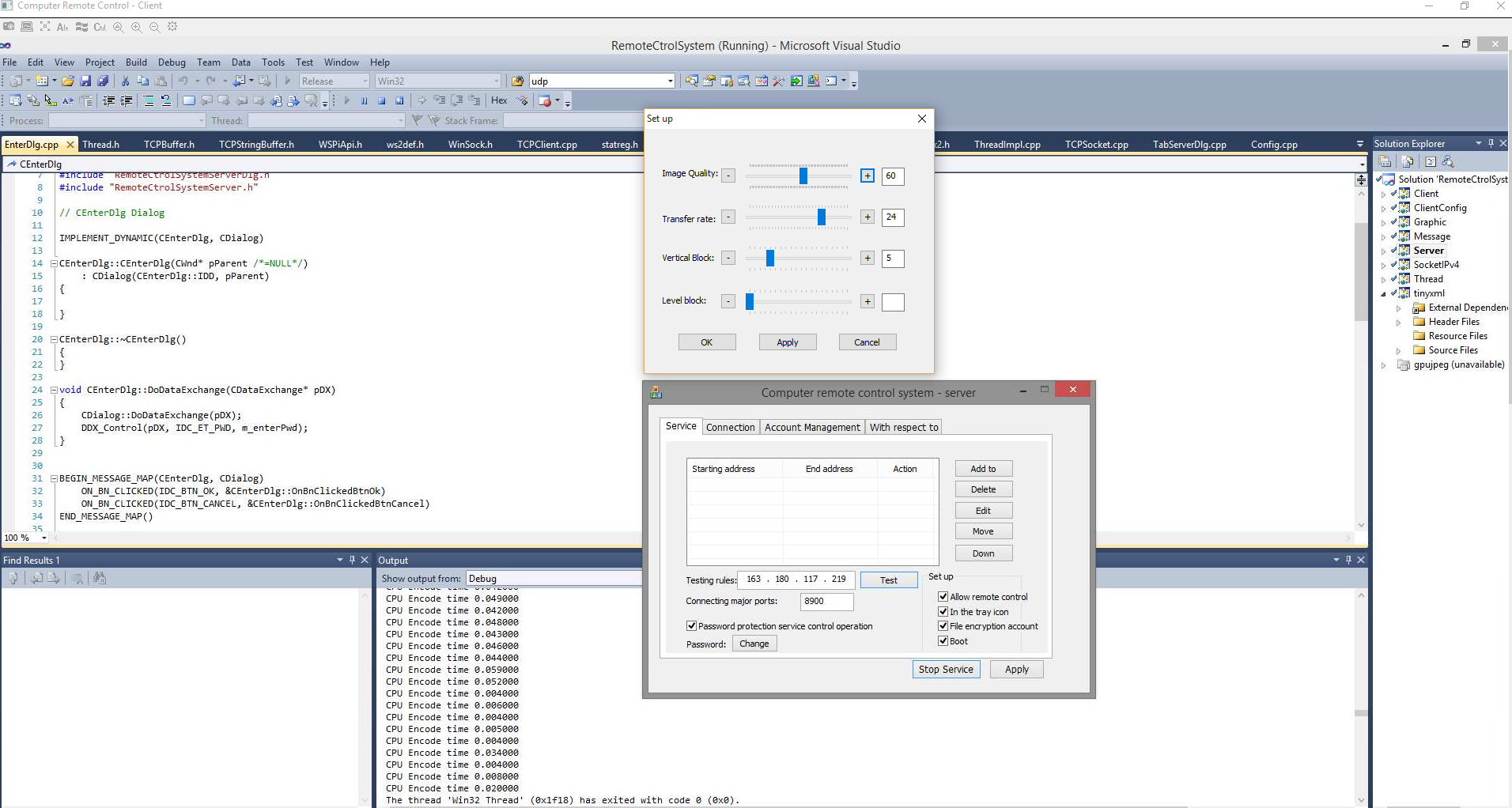}}{b) Screen Desktop 
	}
	
	\caption{System Interface in the cloud server and client side when delivering the video data and the screen desktop}
	\label{fig:input_trsnsmission_data}
\end{figure}

We built a testbed to carry out performance evaluations of the proposed system. Specifically, in the cloud environment, as a visualization platform we used an HP Server (4-core Intel Xeon CPU E3-1220 V2 running at 3.10 GHz, 32 GB RAM) with CentOS 6.5 and OpenStack (Icehouse version) \cite{openstack} installed. Virtual machines (VMs) were installed with lightweight Windows. The VMs ran our server software, which was implemented in C/C++ using Microsoft Visual Studio 2010. We considered low-motion (Screen-Desktop) and high-motion (Korean Movie) videos for transmission Figure~\ref{fig:input_trsnsmission_data}. For simplicity, we set up the environment to run two VMs and measured the frame loss rate during the transmission in two cases: with and without applying the QoS support system, for both video types. It was assumed that the client connected to the service side was a PC device, and the screen resolution and frame rate of all video sequences were SVGA ($800\times600$) and 24 fps, respectively. Thus, the system predicted the best $\mu$ for each frame of desktop image data delivered.
\begin{figure}[!h]
	\includegraphics[width=1\textwidth]{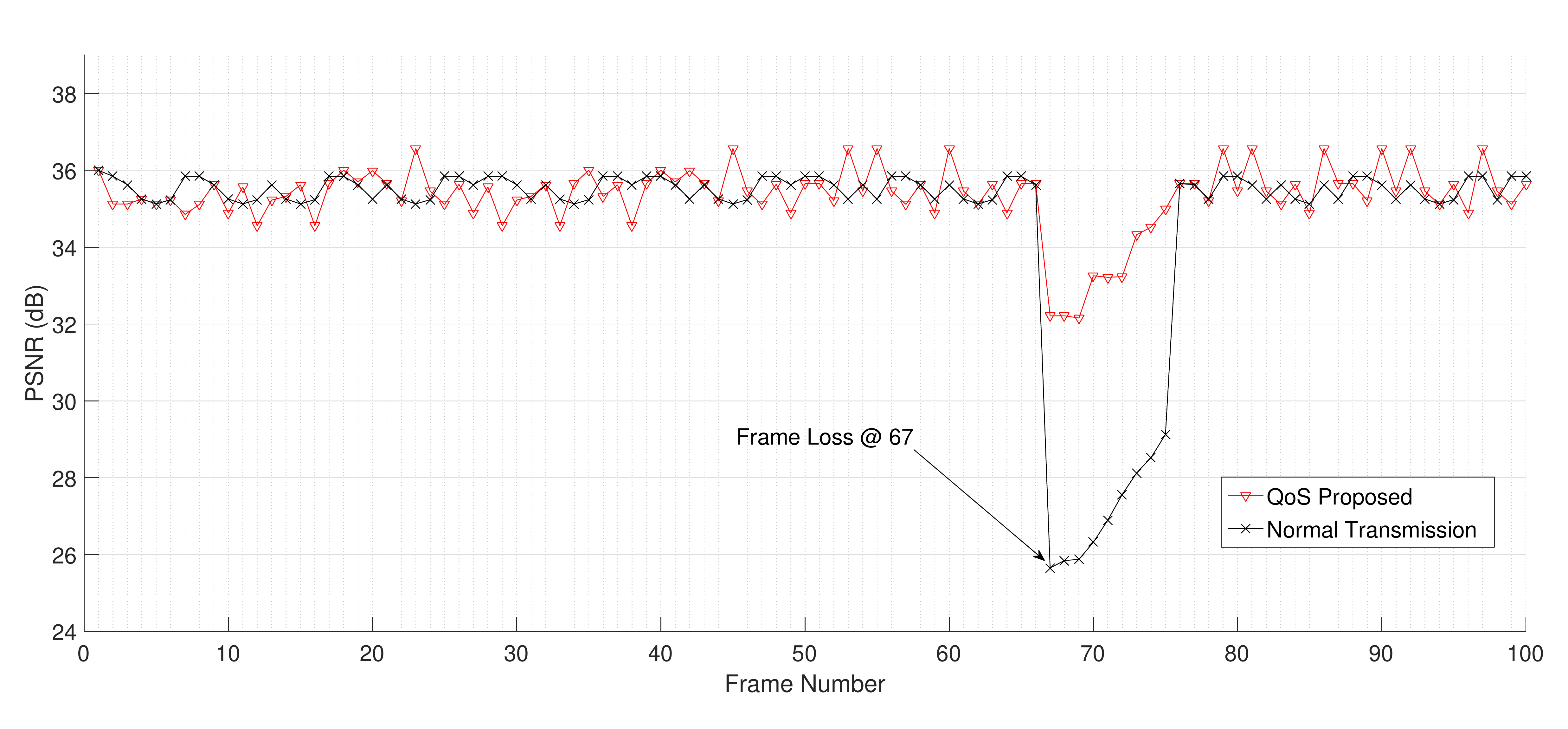}
	\caption{PSNR of low-motion video, which encounters a single lost frame at frame 67, with and without QoS control}
	\label{fig_frameloss_low}       
\end{figure}

\begin{figure}[!h]
	\includegraphics[width=1\textwidth]{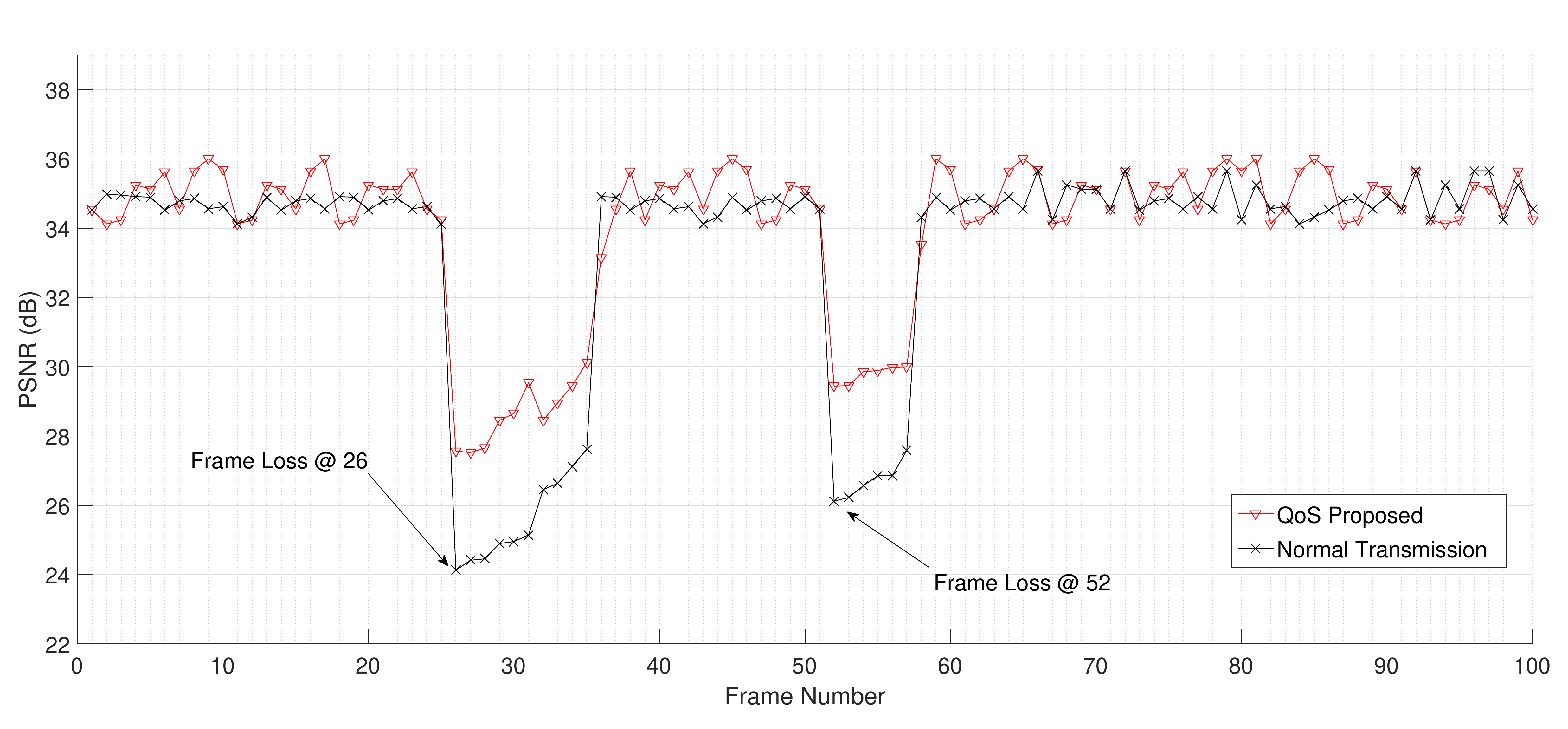}
	\caption{PSNR of high-motion video, which encounters single lost frames at frames 26 and 52, with and without QoS control}
	\label{fig_frameloss_high}       
\end{figure}
\begin{figure}[!h]
	\captionsetup[subfigure]{labelformat=empty}
	\centering
	\footnotesize
	
	\stackunder[5pt]{\includegraphics[width=5.5cm,height=4cm]{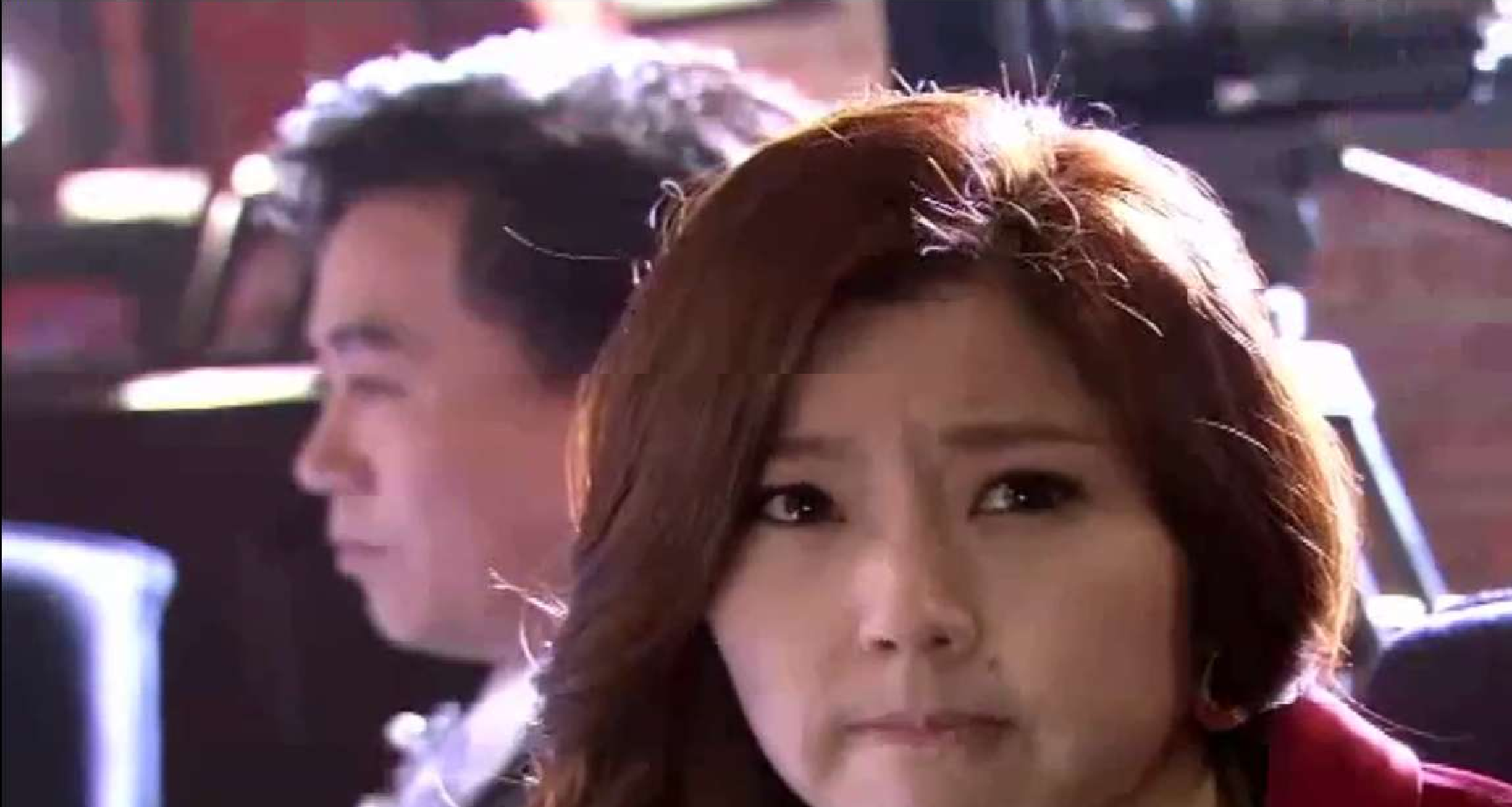}}{a) Normal Transmission}
	\stackunder[5pt]{\includegraphics[width=5.5cm,height=4cm]{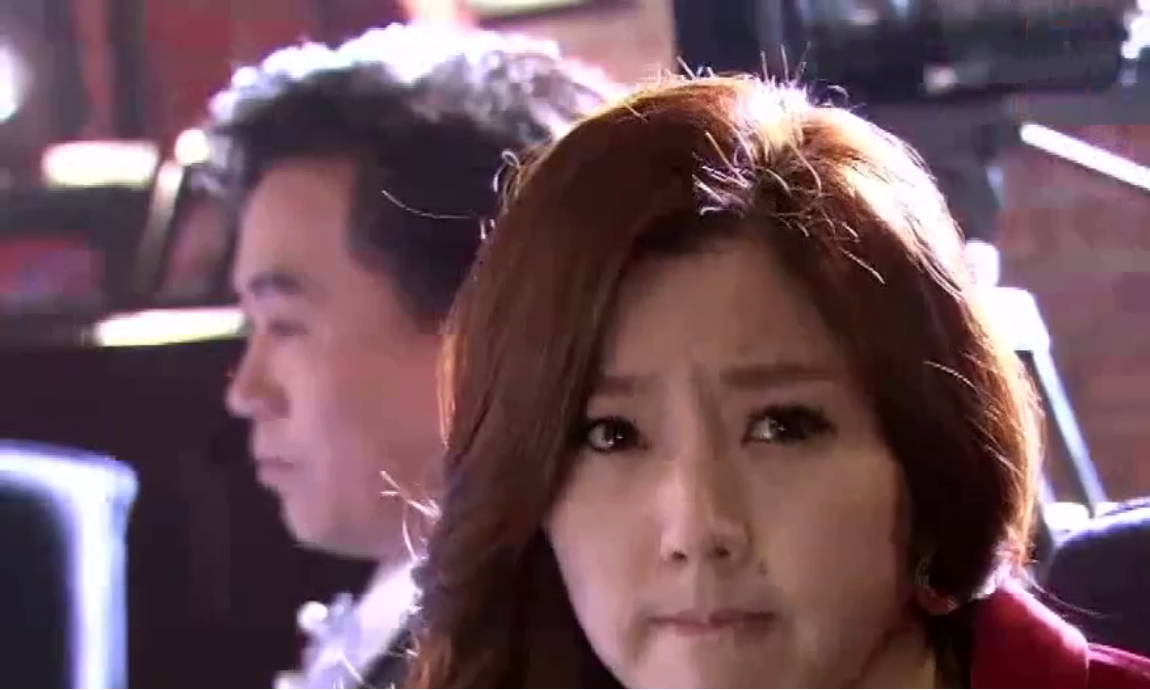}}{b) Transmission with QoS Support}
	\stackunder[5pt]{\includegraphics[width=5.5cm,height=4cm]{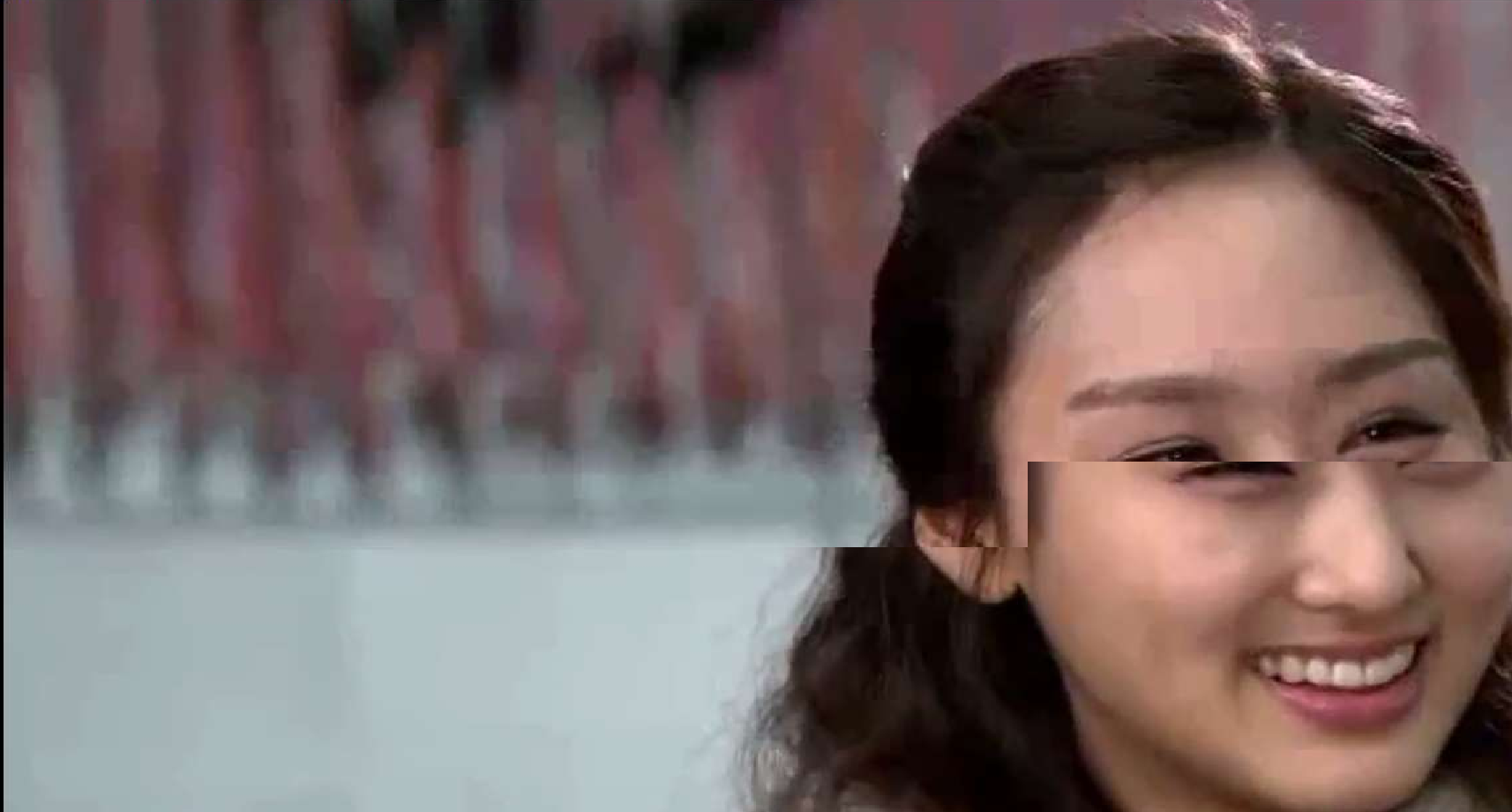}}{a) Normal Transmission}
	\stackunder[5pt]{\includegraphics[width=5.5cm,height=4cm]{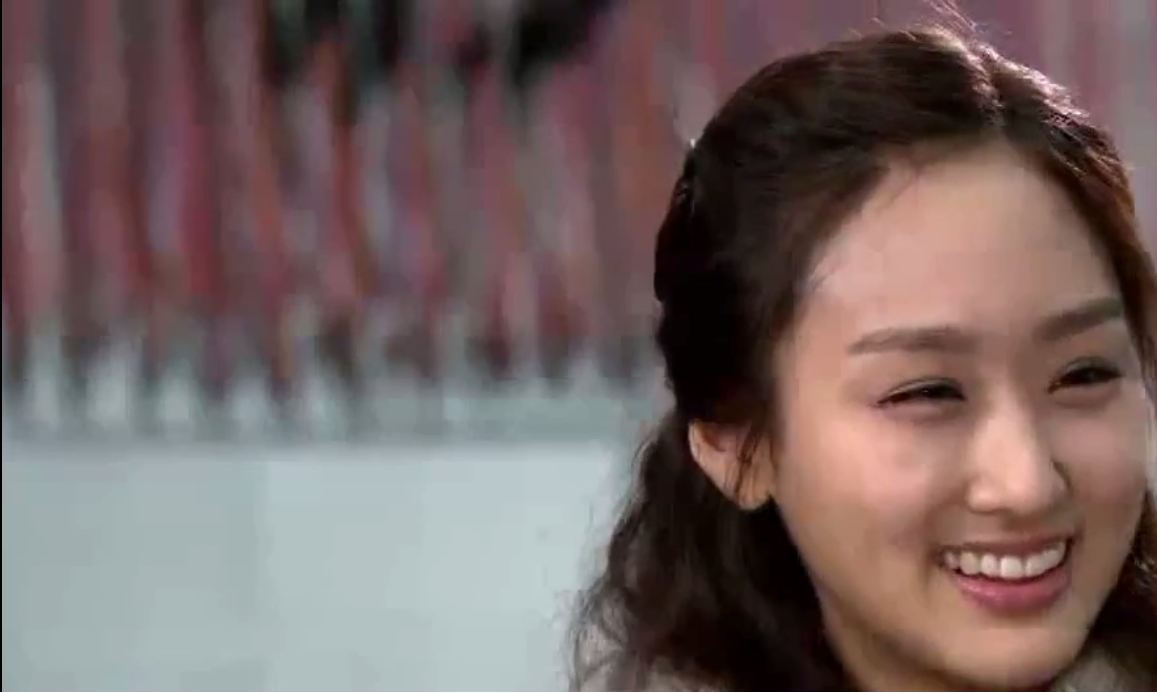}}{b) Transmission with QoS Support}		
	\caption{Comparison of normal transmission and transmission with QoS support}
	\label{fig:input_loss}
\end{figure}

Figure~\ref{fig_frameloss_low} and~\ref{fig_frameloss_high} illustrate the changes that occur in the PSNR when a random single losses occurs during normal transmission and during transmission when the QoS support system is applied. These figures clearly demonstrate the distortion propagation for each type of video. Obviously, QoS policies can help the system to automatically adapt to the network condition. Specifically, the DDEM determines the suitable $\mu$ value. Then, the system automatically adjusts the size of the image to adapt the network status, mitigating the effects of the frame loss occurring at frame 63 in the low-motion video and at frames 26 and 52 in the high-motion videos while ensuring the highest possible quality. 

As shown in Figure~\ref{fig:input_loss} something mismatch for a slice area in the whole picture. This accounts for mismatch that some performed area in the picture is decoded by copying the data from previous key frame, but which cannot match the other part because the right reference data were lost.

\section{Conclusions}
This paper presented a new architecture with flexible QoS control for Virtual Desktop Infrastructure. The proposed architecture includes a novel compression method for 2D images, based on k-means clustering. This method is applied to significantly reduce the size of video data  transmitted while ensuring the highest quality possible. Additionally, to improve users’ quality of experience (QoE) by considering network conditions, we included a model to estimate the most suitable decision for streaming policies, based on the analysis of historical data using linear regression modeling. Through simulations with a real-world dataset, we show the experimental as well as the performance of QoS system that our approach outperforms previously reported methods. The present work is expected to open an important avenue for future related research.

\clearpage
\appendix
\section{Compression Results}

\begin{figure}[!h]
	\captionsetup[subfigure]{labelformat=empty}
	\centering
	\footnotesize
	
	\stackunder[5pt]{\includegraphics[width=1.8cm,height=1.7cm]{butterfly_zebra.jpg}}{Butterfly}
	\stackunder[5pt]{\includegraphics[width=1.8cm,height=1.7cm]{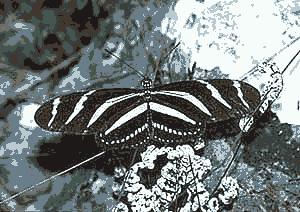}}{$\mu$ = 8}
	\stackunder[5pt]{\includegraphics[width=1.8cm,height=1.7cm]{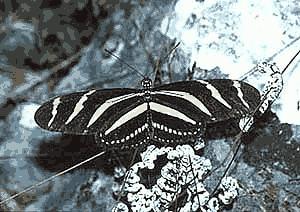}}{$\mu$ = 16}
	\stackunder[5pt]{\includegraphics[width=1.8cm,height=1.7cm]{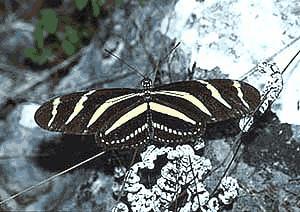}}{$\mu$ = 32}
	\stackunder[5pt]{\includegraphics[width=1.8cm,height=1.7cm]{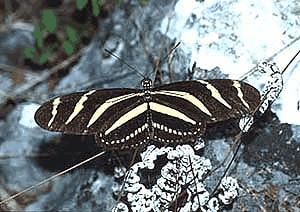}}{$\mu$ = 64}
	\stackunder[5pt]{\includegraphics[width=1.8cm,height=1.7cm]{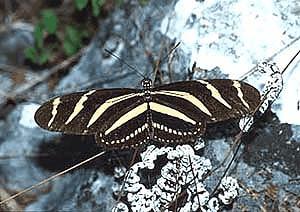}}{$\mu$ = 128}
	
	\stackunder[5pt]{\includegraphics[width=1.8cm,height=1.7cm]{street_boston.jpg}}{Street}
	\stackunder[5pt]{\includegraphics[width=1.8cm,height=1.7cm]{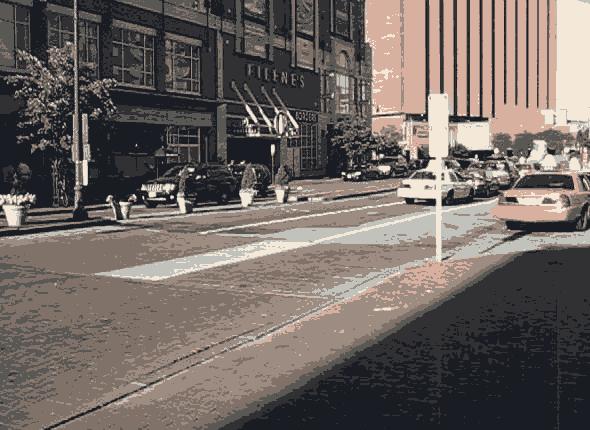}}{$\mu$ = 8}
	\stackunder[5pt]{\includegraphics[width=1.8cm,height=1.7cm]{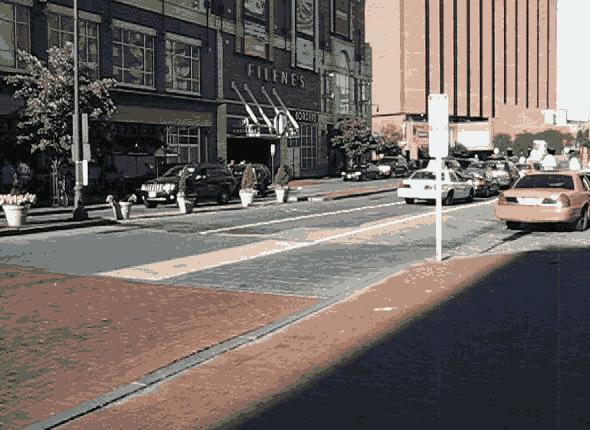}}{$\mu$ = 16}
	\stackunder[5pt]{\includegraphics[width=1.8cm,height=1.7cm]{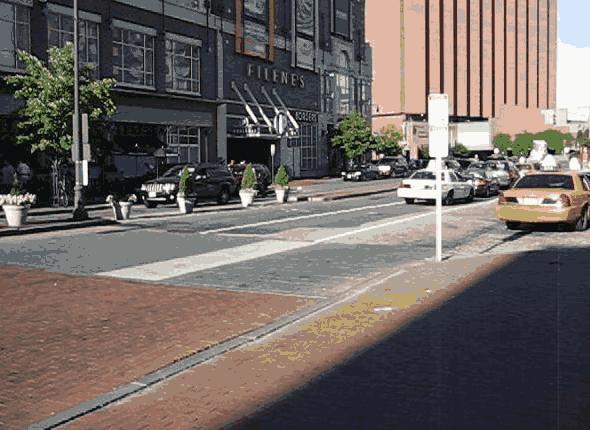}}{$\mu$ = 32}
	\stackunder[5pt]{\includegraphics[width=1.8cm,height=1.7cm]{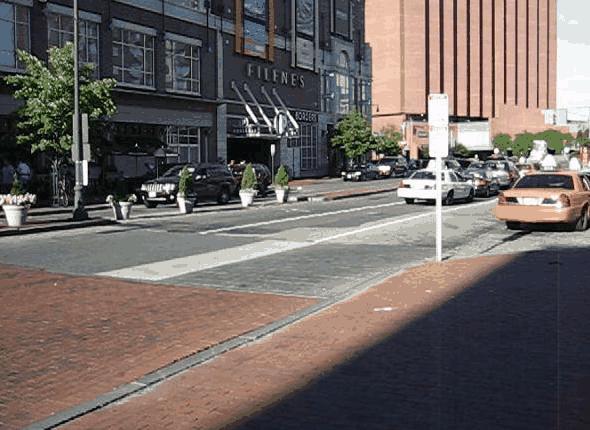}}{$\mu$ = 64}
	\stackunder[5pt]{\includegraphics[width=1.8cm,height=1.7cm]{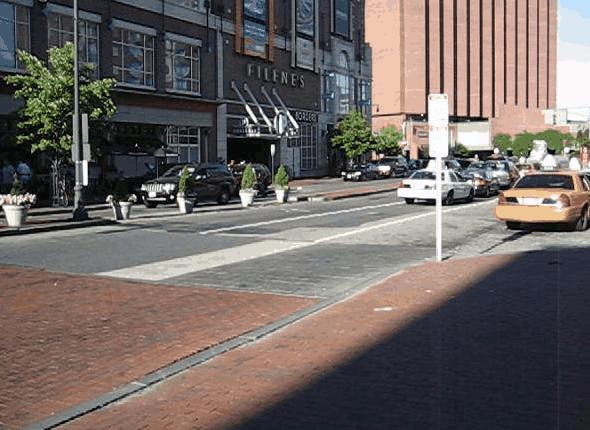}}{$\mu$ = 128}
	
	\stackunder[5pt]{\includegraphics[width=1.8cm,height=1.7cm]{conference_room.jpg}}{Room}
	\stackunder[5pt]{\includegraphics[width=1.8cm,height=1.7cm]{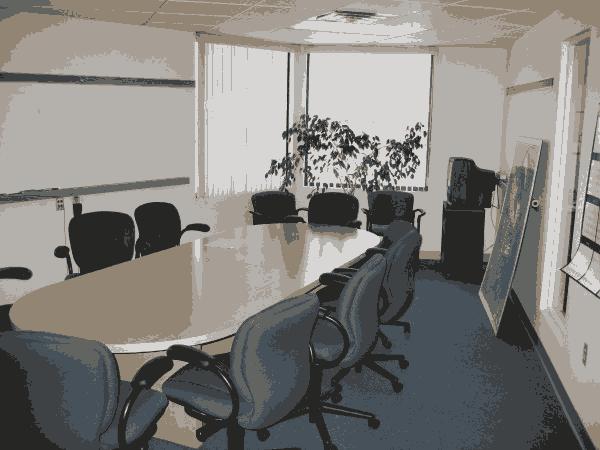}}{$\mu$ = 8}
	\stackunder[5pt]{\includegraphics[width=1.8cm,height=1.7cm]{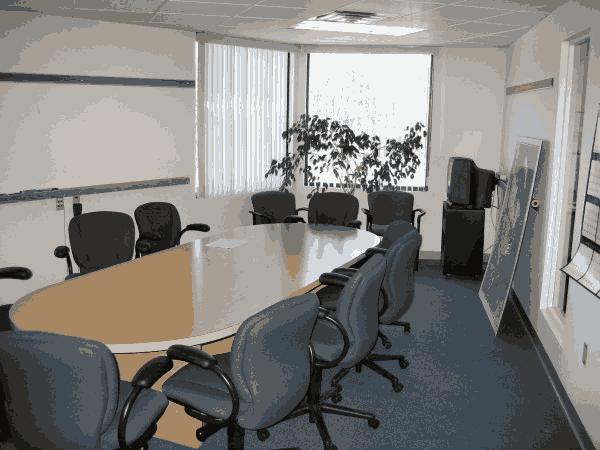}}{$\mu$ = 16}
	\stackunder[5pt]{\includegraphics[width=1.8cm,height=1.7cm]{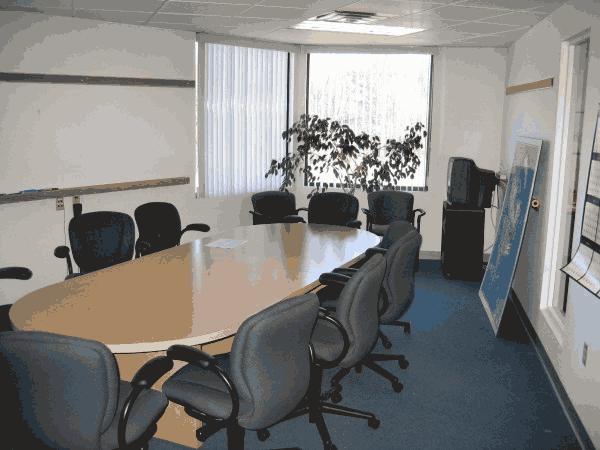}}{$\mu$ = 32}
	\stackunder[5pt]{\includegraphics[width=1.8cm,height=1.7cm]{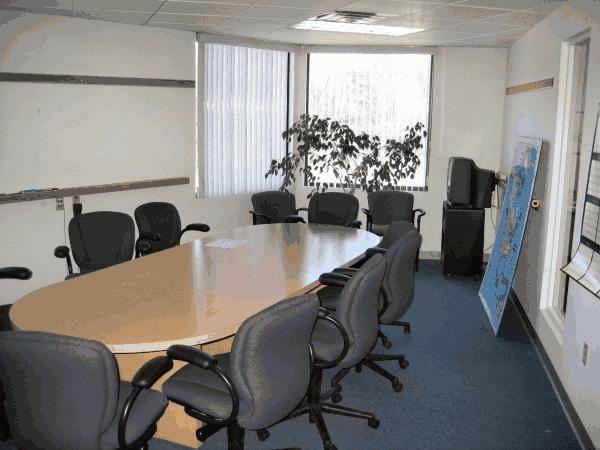}}{$\mu$ = 64}
	\stackunder[5pt]{\includegraphics[width=1.8cm,height=1.7cm]{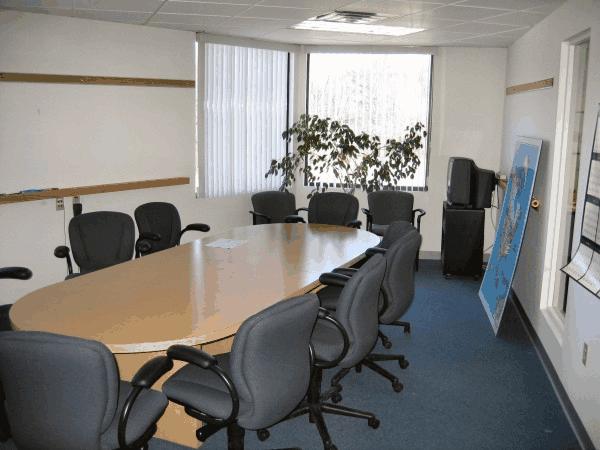}}{$\mu$ = 128}
	
	\stackunder[5pt]{\includegraphics[width=1.8cm,height=1.7cm]{airport.jpg}}{Airport}
	\stackunder[5pt]{\includegraphics[width=1.8cm,height=1.7cm]{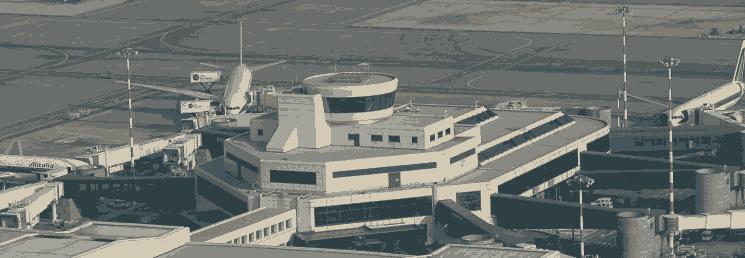}}{$\mu$ = 8}
	\stackunder[5pt]{\includegraphics[width=1.8cm,height=1.7cm]{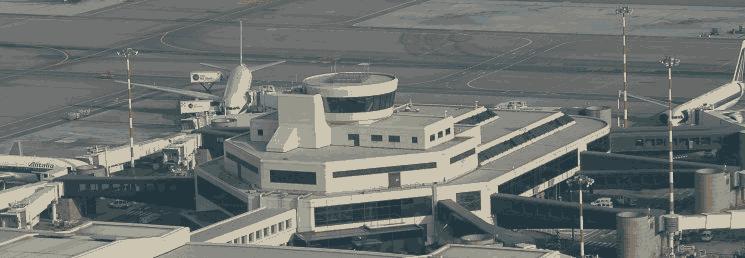}}{$\mu$ = 16}
	\stackunder[5pt]{\includegraphics[width=1.8cm,height=1.7cm]{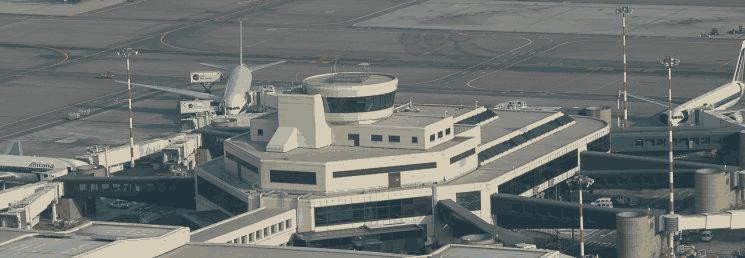}}{$\mu$ = 32}
	\stackunder[5pt]{\includegraphics[width=1.8cm,height=1.7cm]{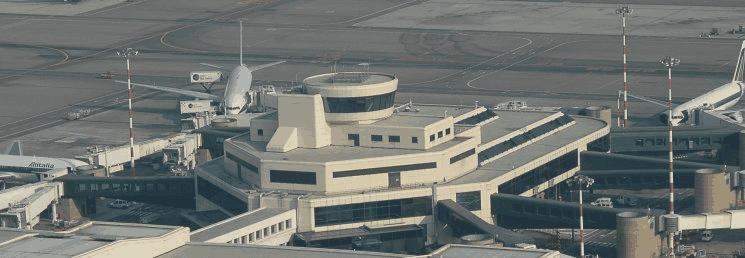}}{$\mu$ = 64}
	\stackunder[5pt]{\includegraphics[width=1.8cm,height=1.7cm]{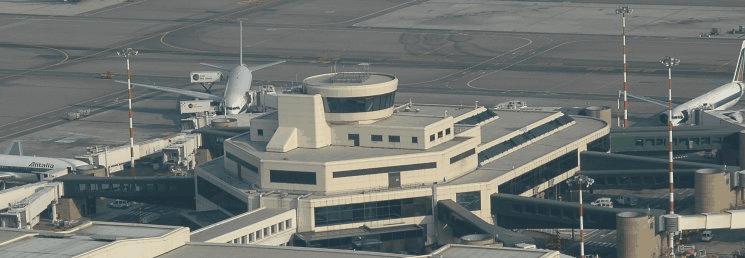}}{$\mu$ = 128}
	
	\stackunder[5pt]{\includegraphics[width=1.8cm,height=1.7cm]{outdoor_oxford.jpg}}{Outdoor}
	\stackunder[5pt]{\includegraphics[width=1.8cm,height=1.7cm]{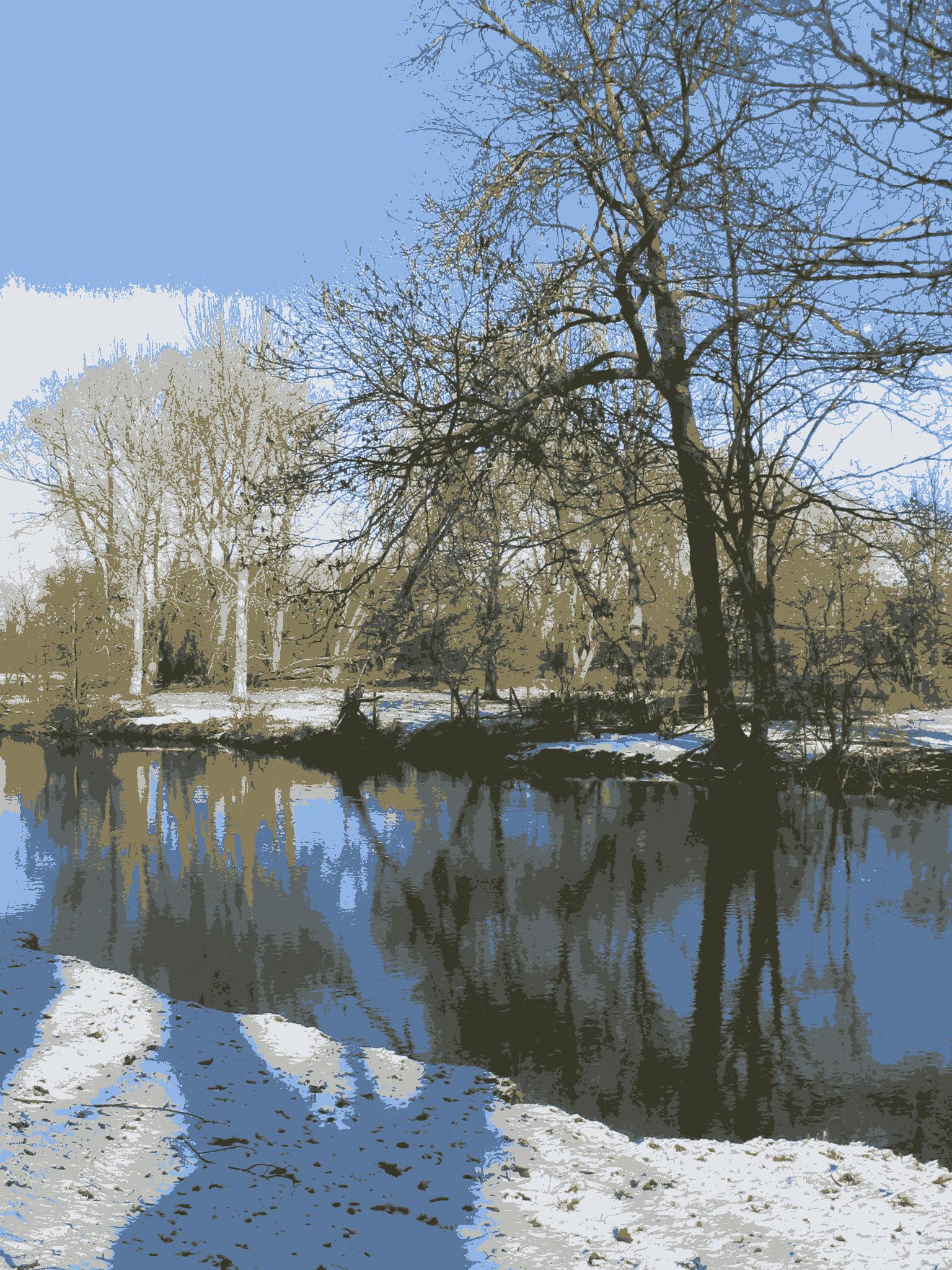}}{$\mu$ = 8}
	\stackunder[5pt]{\includegraphics[width=1.8cm,height=1.7cm]{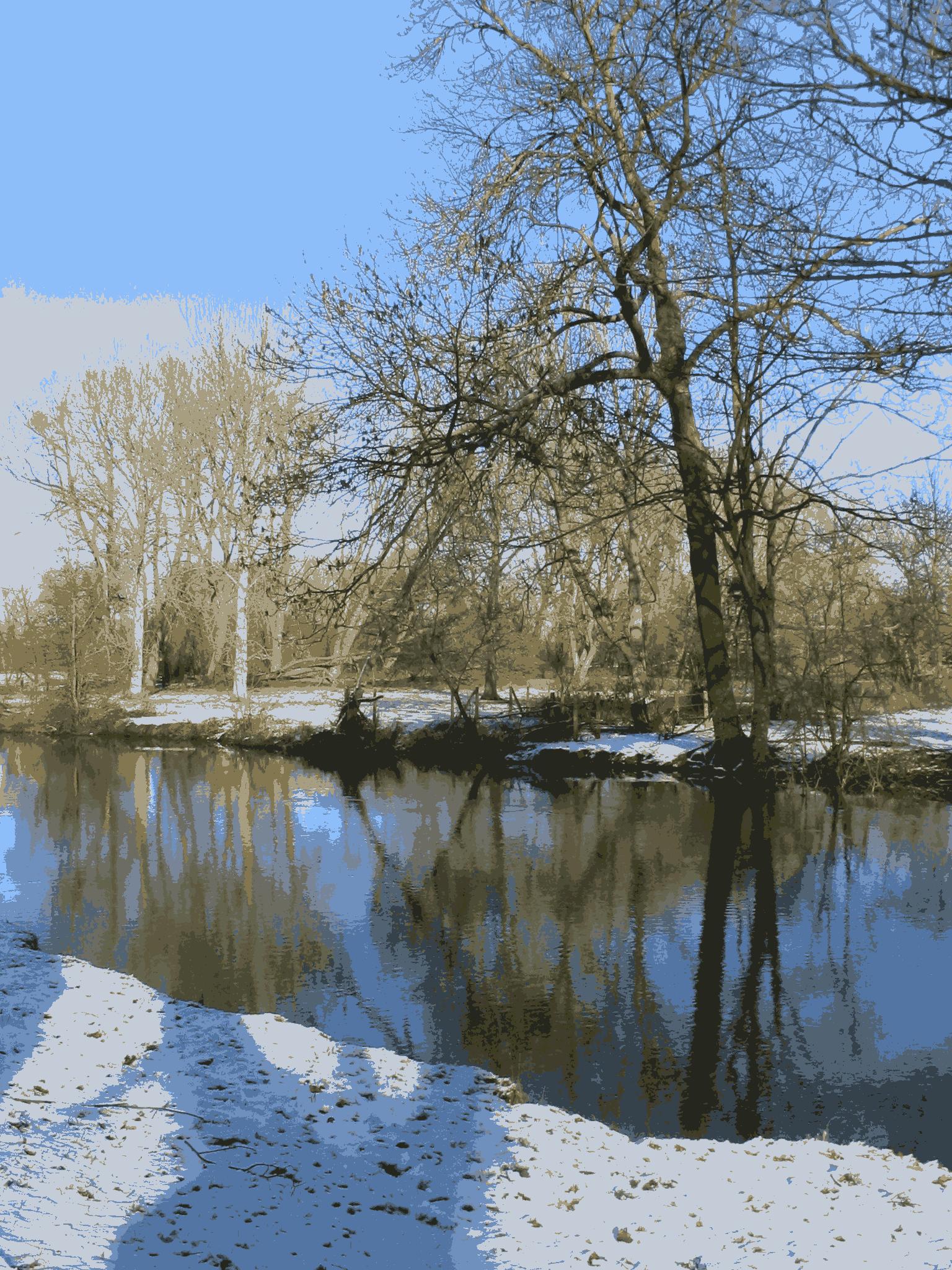}}{$\mu$ = 16}
	\stackunder[5pt]{\includegraphics[width=1.8cm,height=1.7cm]{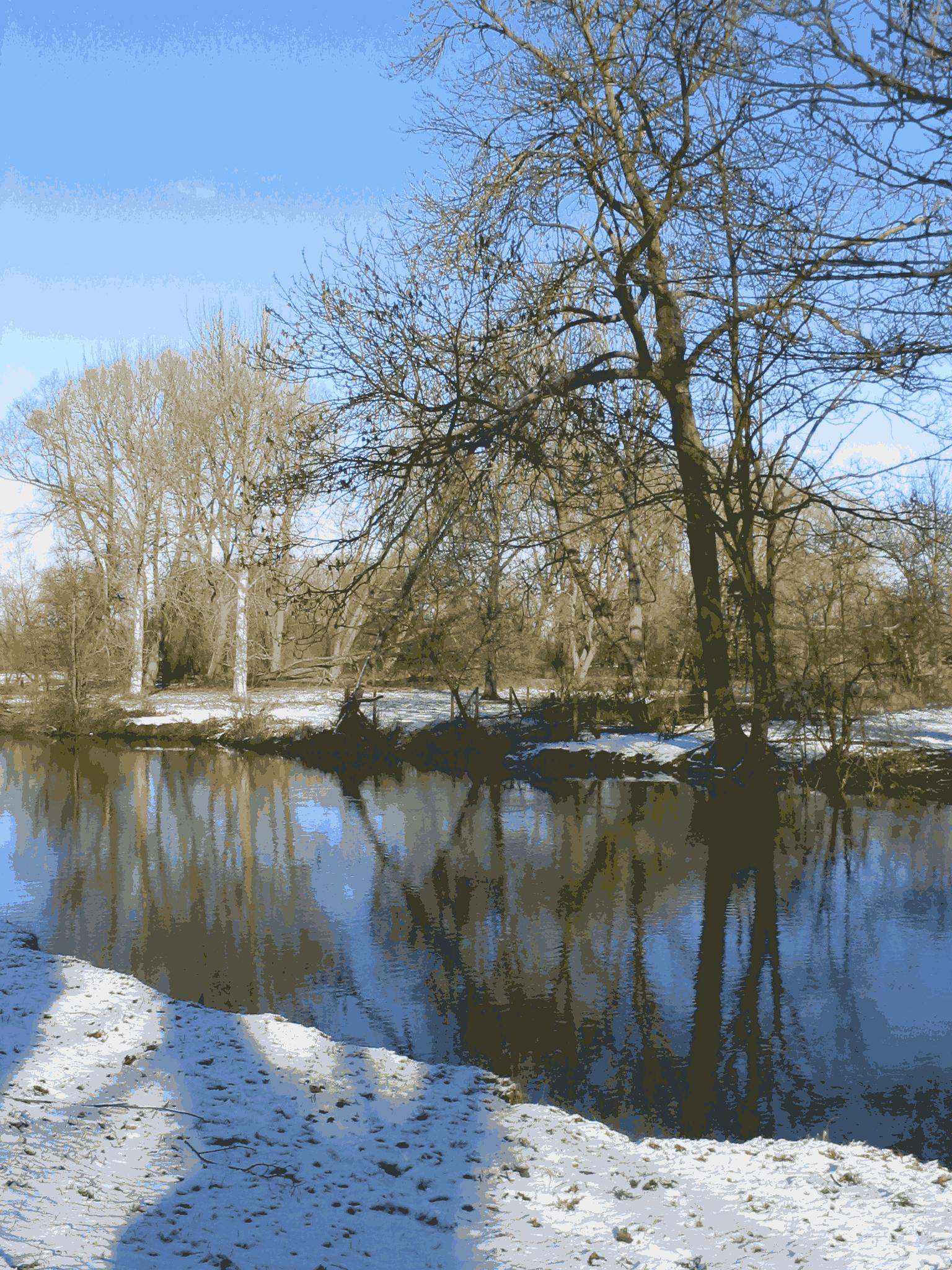}}{$\mu$ = 32}
	\stackunder[5pt]{\includegraphics[width=1.8cm,height=1.7cm]{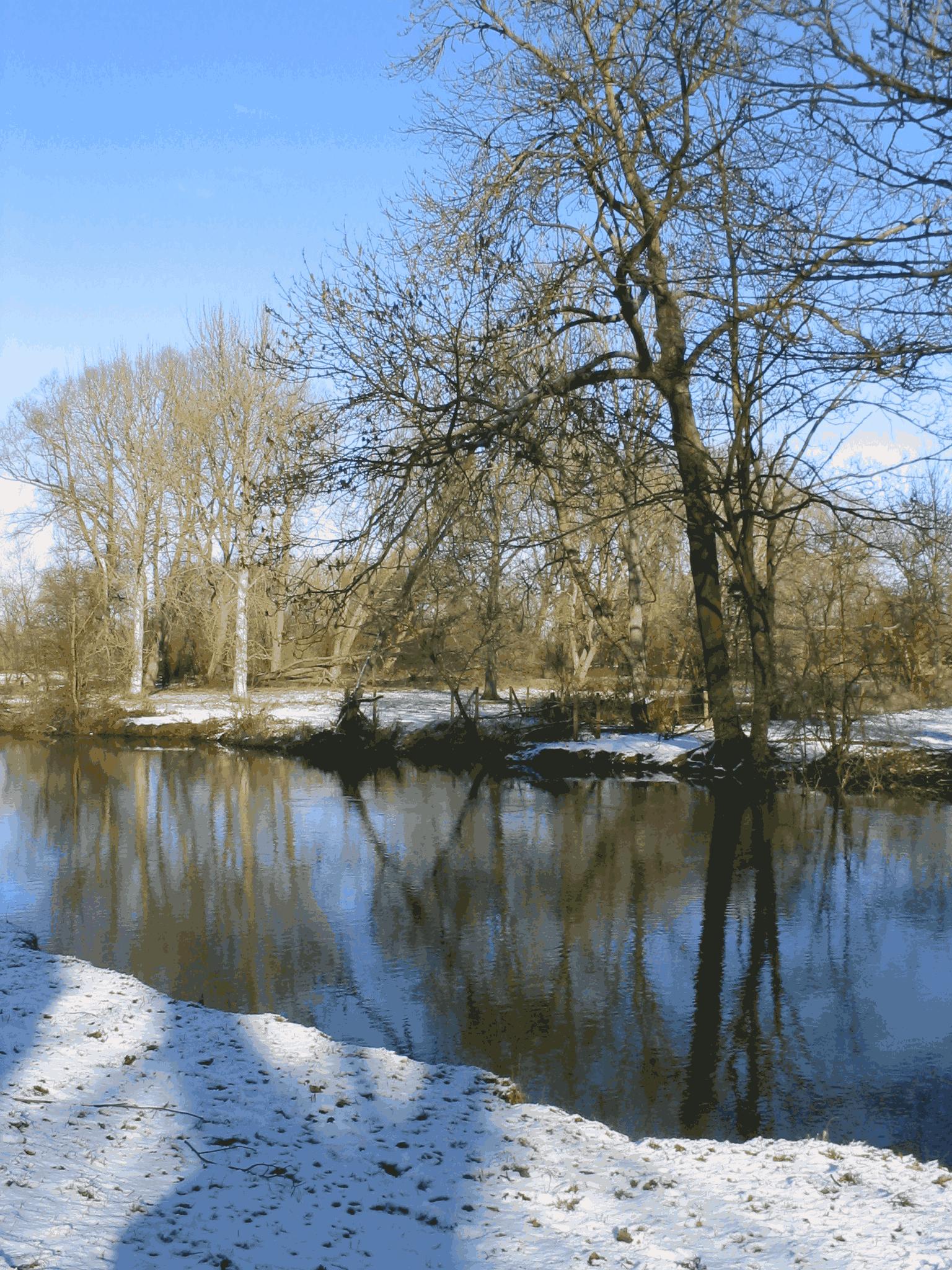}}{$\mu$ = 64}
	\stackunder[5pt]{\includegraphics[width=1.8cm,height=1.7cm]{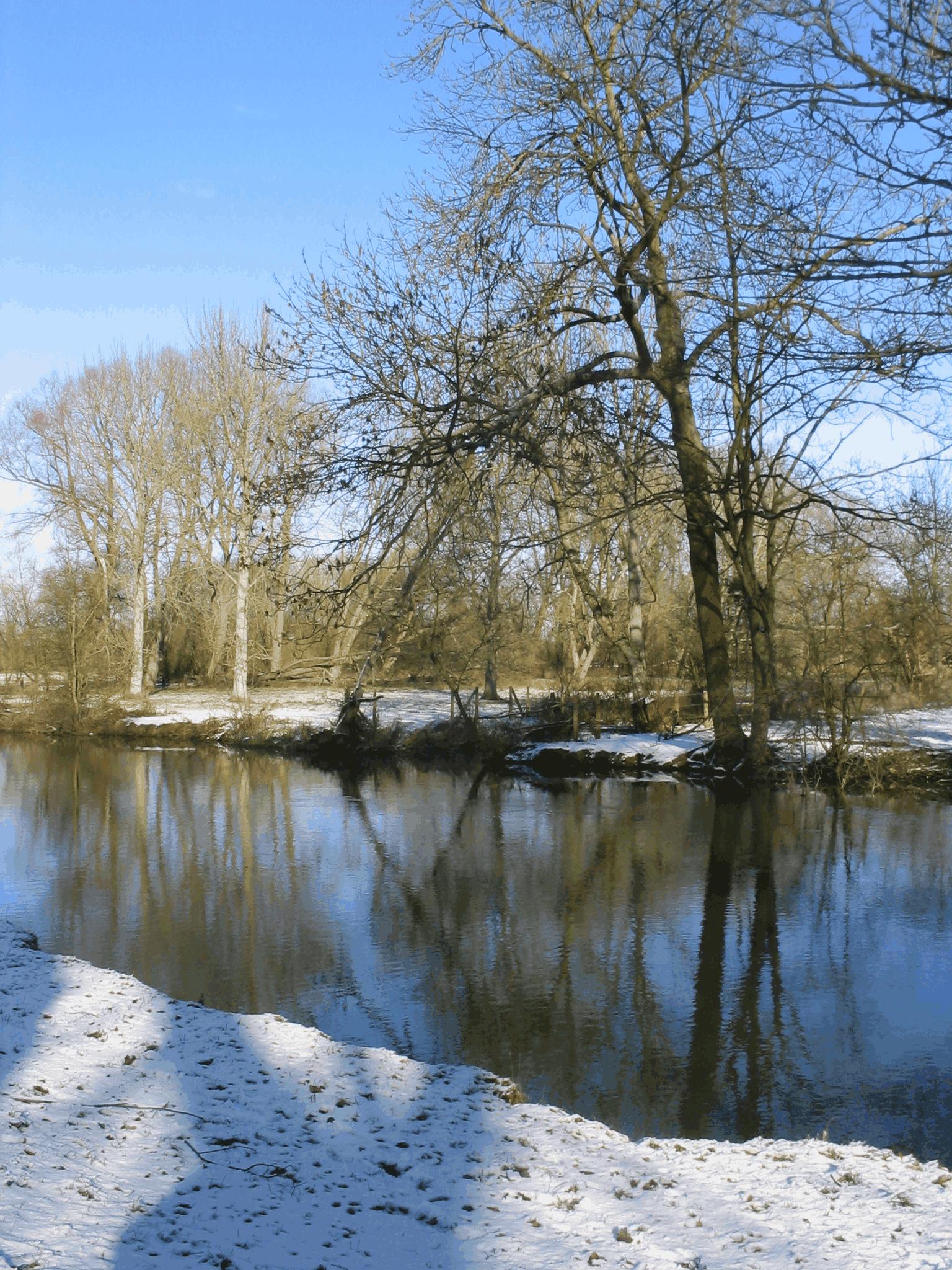}}{$\mu$ = 128}
	
	\stackunder[5pt]{\includegraphics[width=1.8cm,height=1.7cm]{angry_birds.jpg}}{Angry Bird}
	\stackunder[5pt]{\includegraphics[width=1.8cm,height=1.7cm]{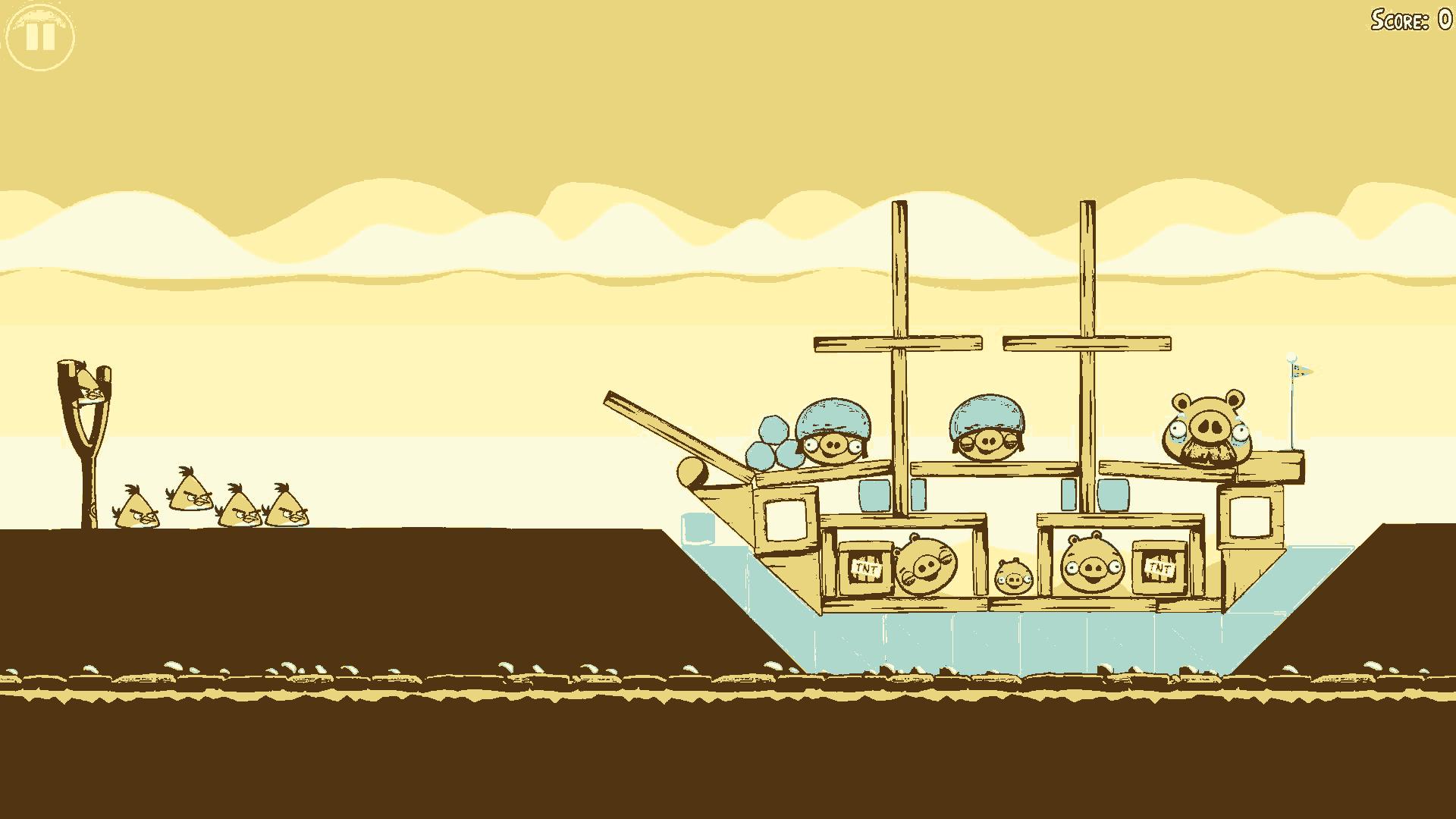}}{$\mu$ = 8}
	\stackunder[5pt]{\includegraphics[width=1.8cm,height=1.7cm]{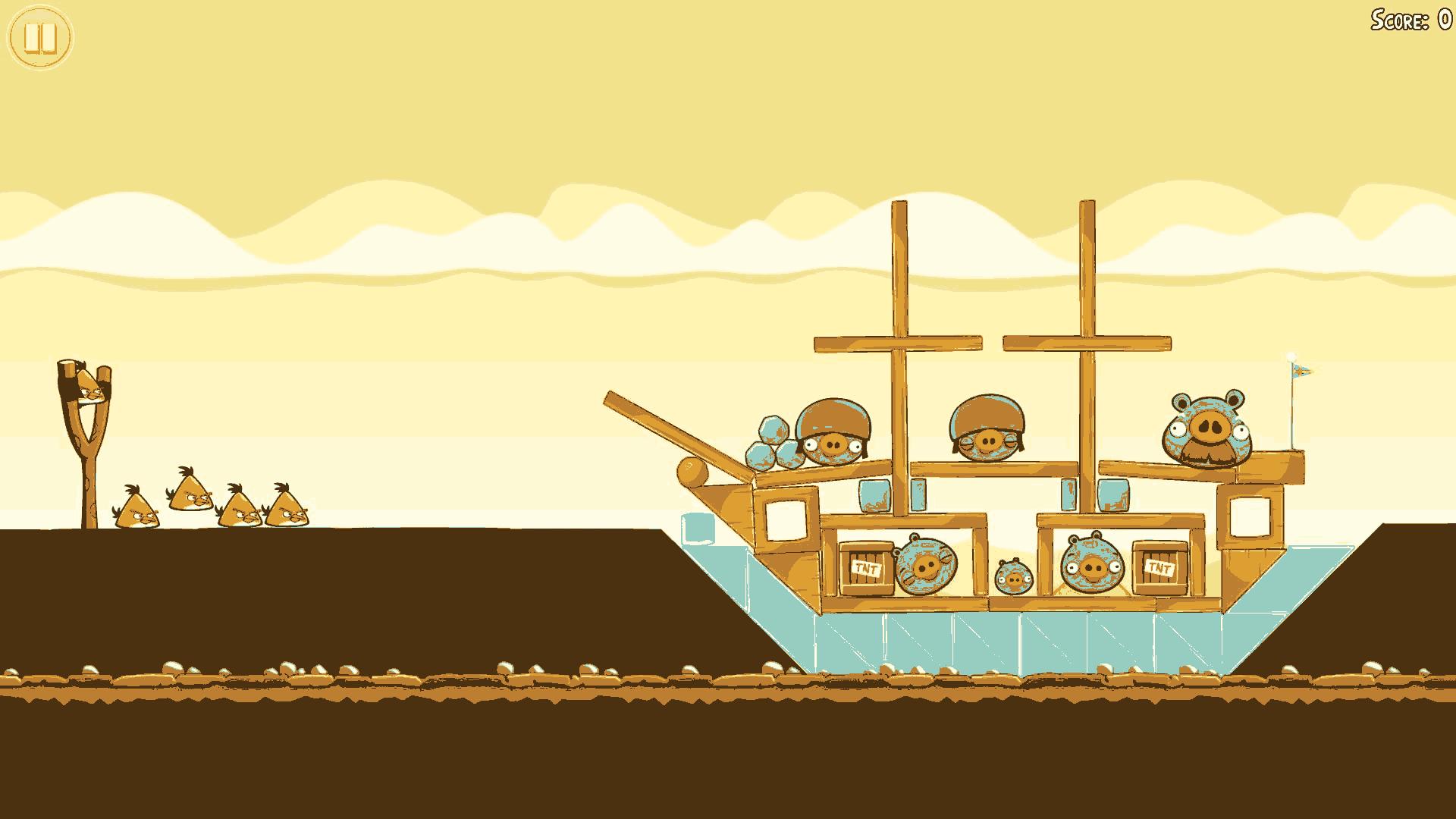}}{$\mu$ = 16}
	\stackunder[5pt]{\includegraphics[width=1.8cm,height=1.7cm]{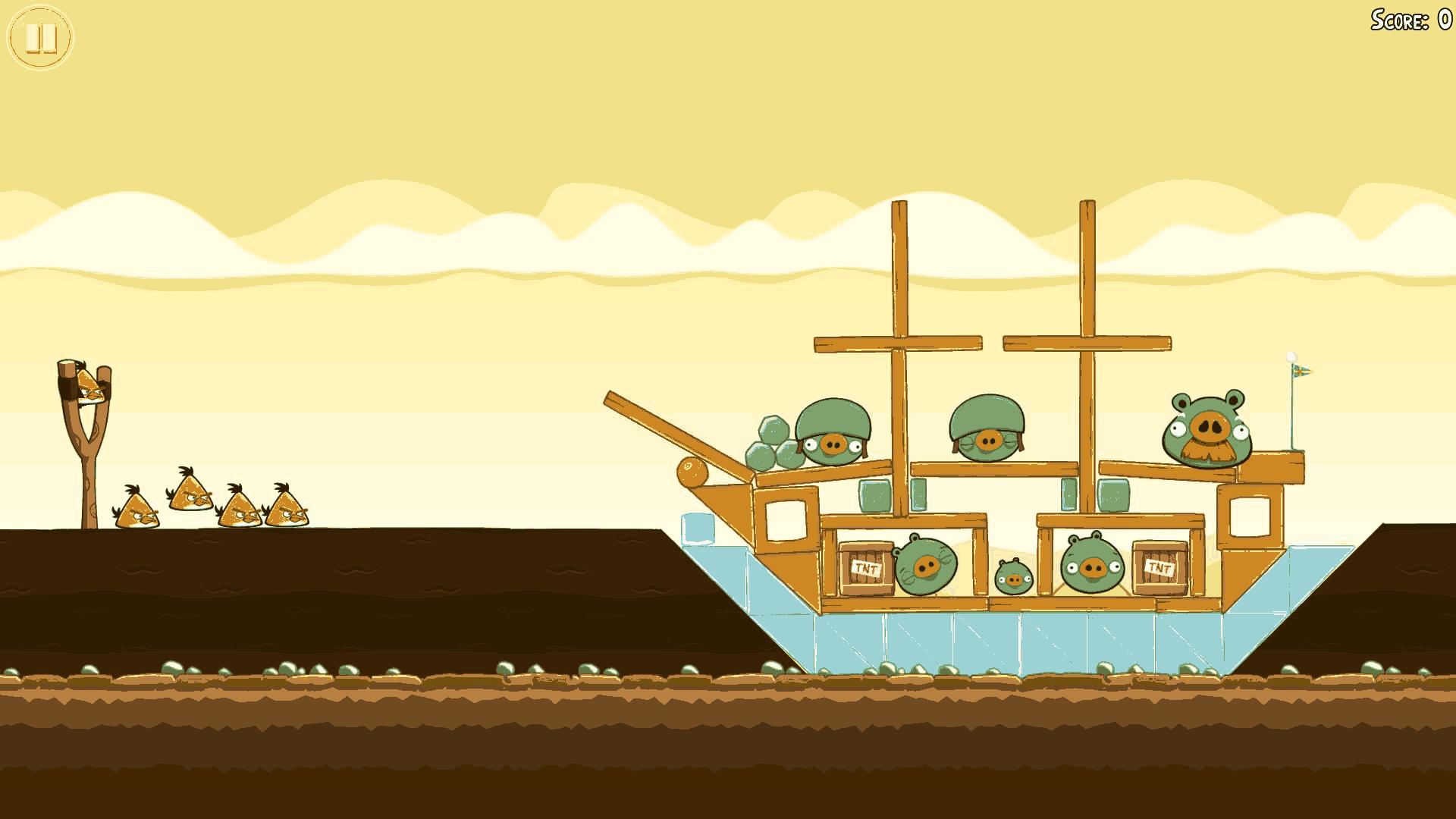}}{$\mu$ = 32}
	\stackunder[5pt]{\includegraphics[width=1.8cm,height=1.7cm]{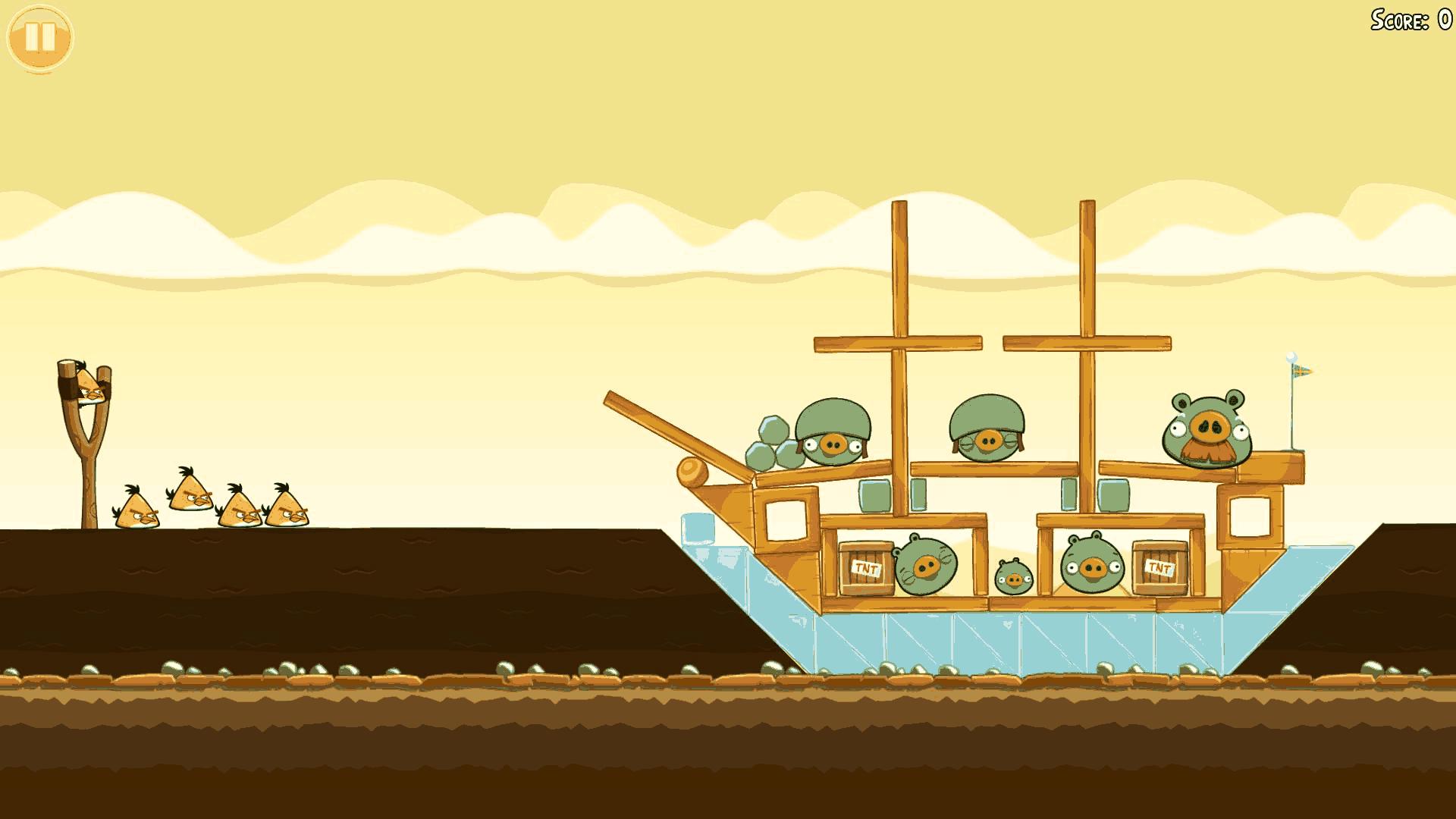}}{$\mu$ = 64}
	\stackunder[5pt]{\includegraphics[width=1.8cm,height=1.7cm]{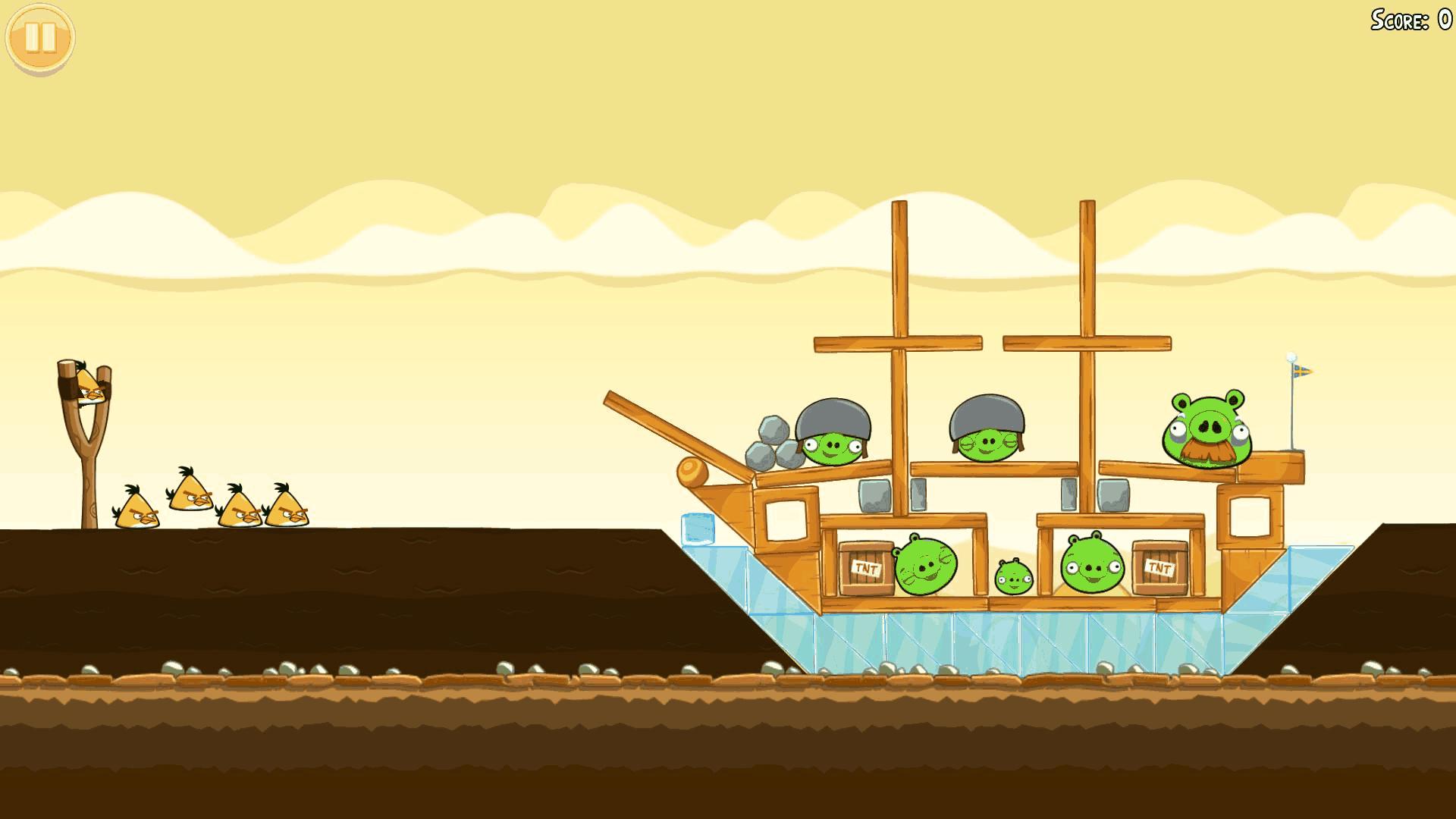}}{$\mu$ = 128}
	
	\stackunder[5pt]{\includegraphics[width=1.8cm,height=1.7cm]{chip_desiner.jpg}}{Tool}
	\stackunder[5pt]{\includegraphics[width=1.8cm,height=1.7cm]{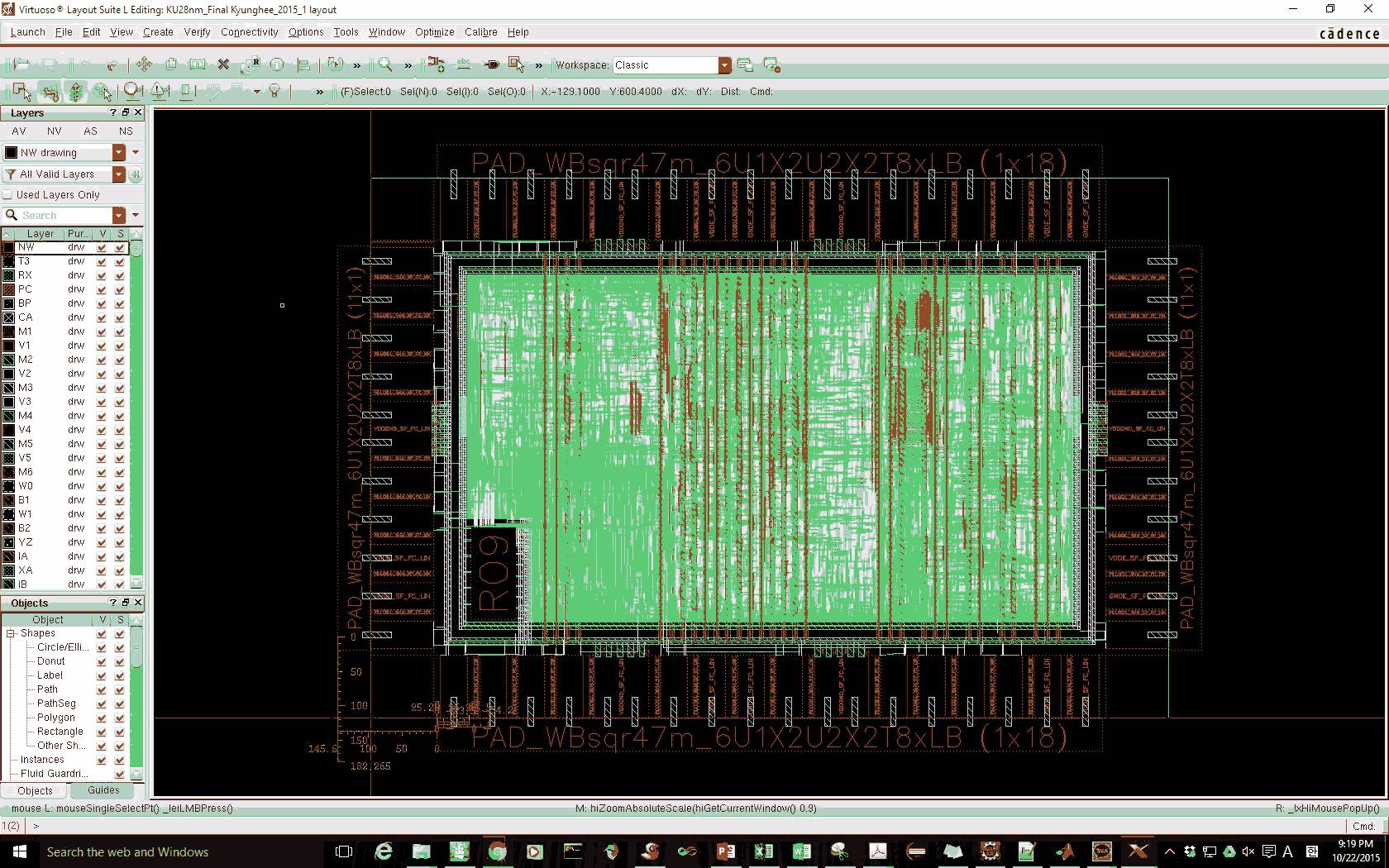}}{$\mu$ = 8}
	\stackunder[5pt]{\includegraphics[width=1.8cm,height=1.7cm]{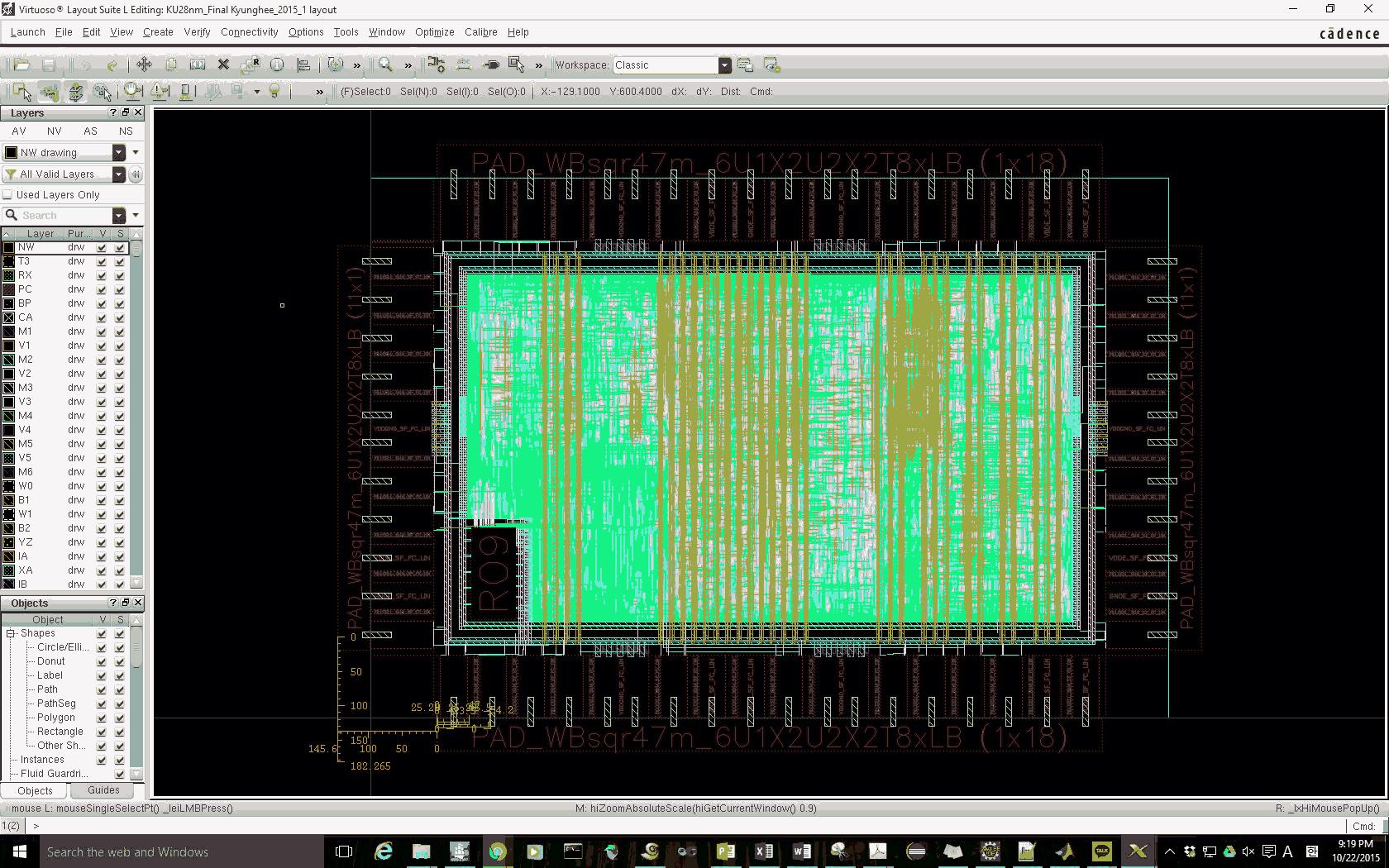}}{$\mu$ = 16}
	\stackunder[5pt]{\includegraphics[width=1.8cm,height=1.7cm]{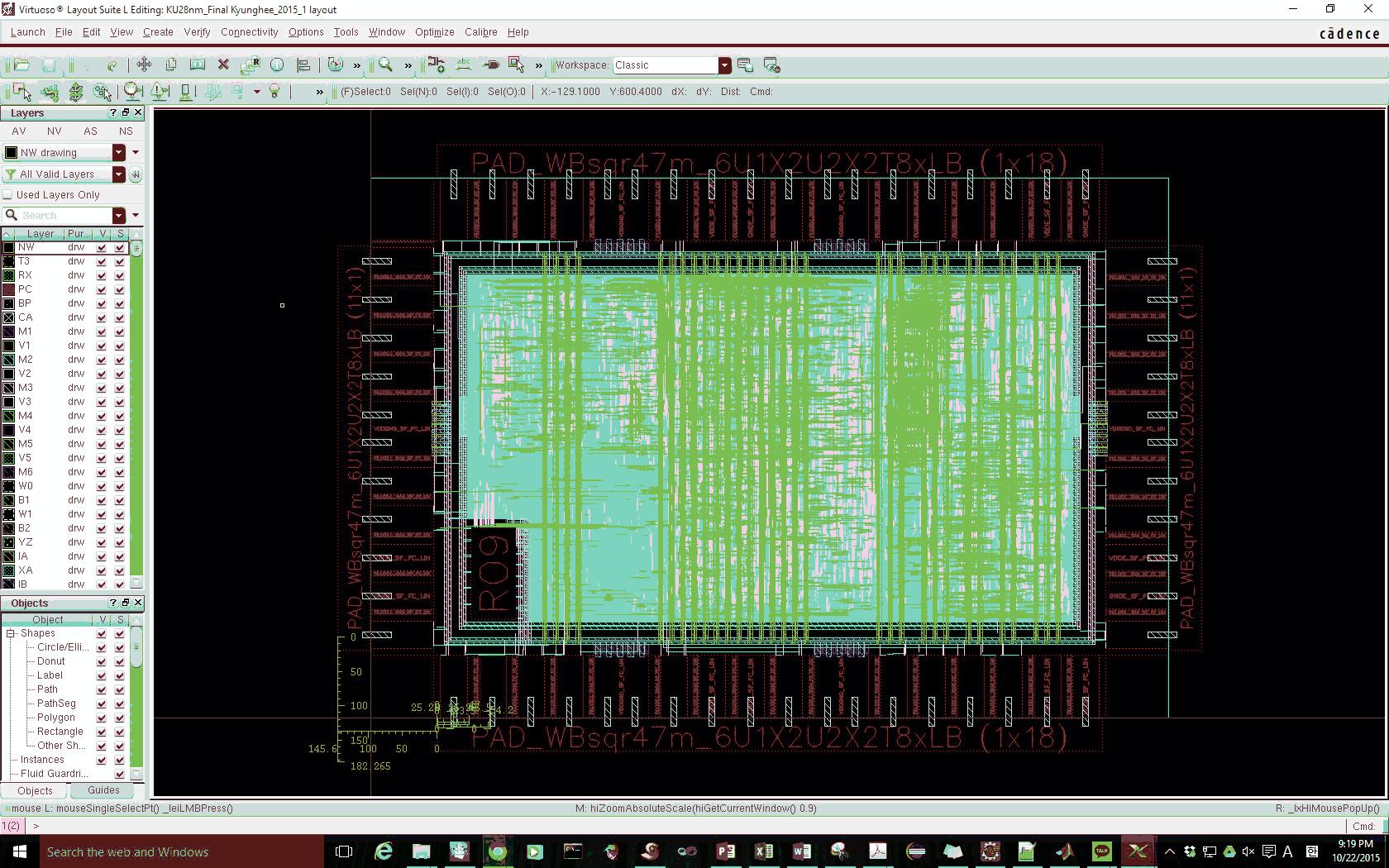}}{$\mu$ = 32}
	\stackunder[5pt]{\includegraphics[width=1.8cm,height=1.7cm]{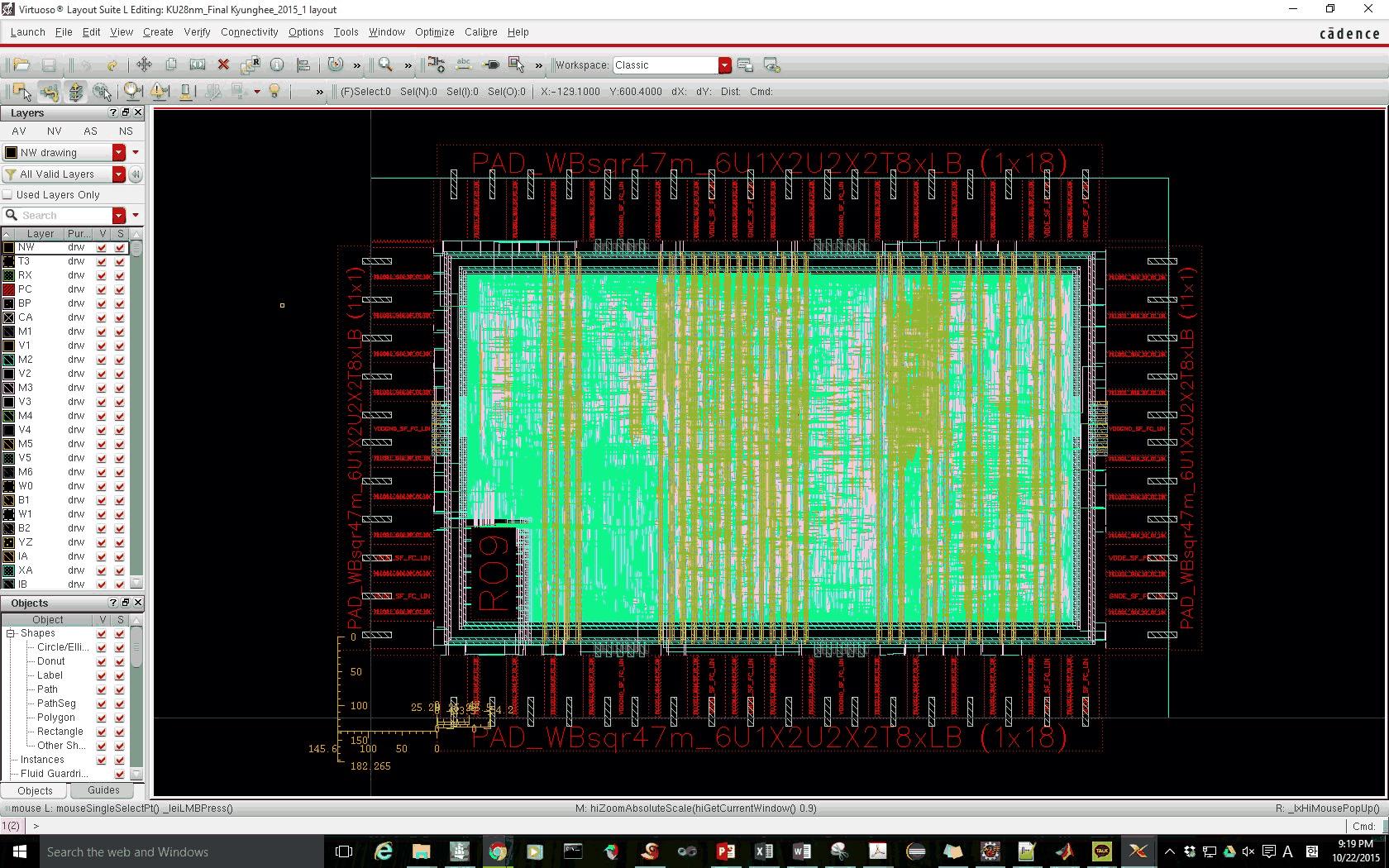}}{$\mu$ = 64}
	\stackunder[5pt]{\includegraphics[width=1.8cm,height=1.7cm]{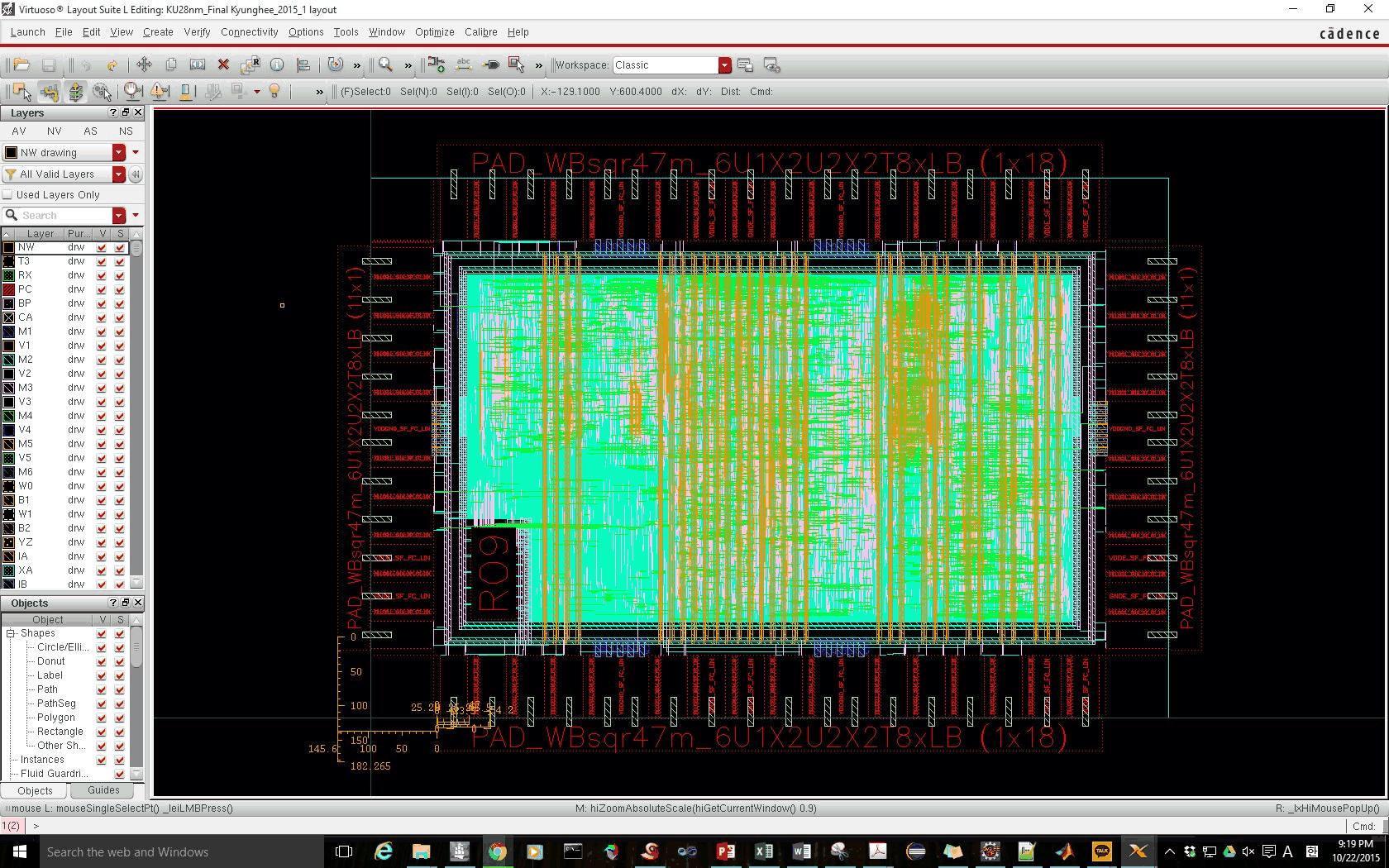}}{$\mu$ = 128}
	
	\caption{Images resulting from compression with various $\mu$}
	\label{fig:output_image}
\end{figure}
\clearpage
\begin{table} [!h]  
	\centering
	\caption{Compression size, PSNR, and compression ratio after the ICM compression phase}
	\label{table_compression_data}       
	\begin{tabular}{ l *{5}{D{.}{.}{4}} }
		
		\toprule
		\textbf{} & \multicolumn{5}{c}{\textbf{1. Zebra Buffterfly (300x212) - 27.4 kB}} \\
		\cmidrule(lr){2-6}
		\textbf{} & \multicolumn{1}{c}{$\mu = 8$} & \multicolumn{1}{c}{$\mu = 16$} & \multicolumn{1}{c}{$\mu = 32$} & \multicolumn{1}{c}{$\mu = 64$} & \multicolumn{1}{c}{$\mu = 128$} \\
		\midrule
		\texttt{Image Size (kB)} &21.7 &20.6 &19.8 &19.8 &19.7 \\
		\texttt{PSNR (dB)} &24.6758 &28.6469 &30.7613 &33.0554 &34.6483 \\		
		\texttt{CS} &1.2626 &1.33 &1.3838 &1.3838 &1.3908 \\	
		
		\toprule
		\textbf{} & \multicolumn{5}{c}{\textbf{2. Boston Street (590x430) - 69.2 kB}} \\
		\cmidrule(lr){2-6}
		\textbf{} & \multicolumn{1}{c}{$\mu = 8$} & \multicolumn{1}{c}{$\mu = 16$} & \multicolumn{1}{c}{$\mu = 32$} & \multicolumn{1}{c}{$\mu = 64$} & \multicolumn{1}{c}{$\mu = 128$} \\
		\midrule
		\texttt{Image Size (kB)} &52.5 &50.2 &49.4 &49.3 &49 \\
		\texttt{PSNR (dB)} &24.8155 &28.0013 &29.8641  &31.9285 &33.9174\\  
		\texttt{CS} &1.3180 &1.3784 &1.40008 &1.4036 &1.4122 \\
		
		\toprule
		\textbf{} & \multicolumn{5}{c}{\textbf{3. Conference Room (600x450) - 61.4 kB}} \\
		\cmidrule(lr){2-6}
		\textbf{} & \multicolumn{1}{c}{$\mu = 8$} & \multicolumn{1}{c}{$\mu = 16$} & \multicolumn{1}{c}{$\mu = 32$} & \multicolumn{1}{c}{$\mu = 64$} & \multicolumn{1}{c}{$\mu = 128$} \\
		\midrule
		\texttt{Image Size (kB)} &36.2 &40.7 &38.7 &39.6 &38 \\
		\texttt{PSNR (dB)} &25.2174 &28.7521 &30.0919  &30.0919 &33.9716\\ 
		\texttt{CS} &1.6961 &1.5085 &1.5865 &1.5505 &1.6157 \\ 
		
		\toprule
		\textbf{} & \multicolumn{5}{c}{\textbf{4. Airport (745x258) - 46.2 kB}} \\
		\cmidrule(lr){2-6}
		\textbf{} & \multicolumn{1}{c}{$\mu = 8$} & \multicolumn{1}{c}{$\mu = 16$} & \multicolumn{1}{c}{$\mu = 32$} & \multicolumn{1}{c}{$\mu = 64$} & \multicolumn{1}{c}{$\mu = 128$} \\
		\midrule
		\texttt{Image Size (kB)} &36.9 &37.1 &37.5 &38.2 &38.6 \\
		\texttt{PSNR (dB)} &30.2741 &31.6957 &33.0064  &34.7103 &35.3905\\ 
		\texttt{CS} &1.2520 &1.2452 &1.2320 &1.2094 &1.1968 \\ 	
		
		\toprule
		\textbf{} & \multicolumn{5}{c}{\textbf{5. Oxford Outdoor (1536x2048) - 1.21 MB}} \\
		\cmidrule(lr){2-6}
		\textbf{} & \multicolumn{1}{c}{$\mu = 8$} & \multicolumn{1}{c}{$\mu = 16$} & \multicolumn{1}{c}{$\mu = 32$} & \multicolumn{1}{c}{$\mu = 64$} & \multicolumn{1}{c}{$\mu = 128$} \\
		\midrule
		\texttt{Image Size (kB)} &653 &612 &619 &603 &589 \\
		\texttt{PSNR (dB)} &23.2208 &25.3275 &28.3941  &31.1382 &32.5477\\ 
		\texttt{CS} &1.8974 &2.0245 &2.0016 &2.0547 &2.1036 \\ 
		
		\toprule
		\textbf{} & \multicolumn{5}{c}{\textbf{6. Chip Designer (1680x1050) - 503 kB}} \\
		\cmidrule(lr){2-6}
		\textbf{} & \multicolumn{1}{c}{$\mu = 8$} & \multicolumn{1}{c}{$\mu = 16$} & \multicolumn{1}{c}{$\mu = 32$} & \multicolumn{1}{c}{$\mu = 64$} & \multicolumn{1}{c}{$\mu = 128$} \\
		\midrule
		\texttt{Image Size (kB)} &374 &338 &339 &355 &358 \\
		\texttt{PSNR (dB)} &20.1001 &22.5293 &20.002  &24.5301 &27.4084\\ 
		\texttt{CS} &1.3449 &1.4881 &1.4837 &1.4169 &1.4050 \\ 
		
		\toprule
		\textbf{} & \multicolumn{5}{c}{\textbf{7. Angry Bird (1920x1080) - 153 kB}} \\
		\cmidrule(lr){2-6}
		\textbf{} & \multicolumn{1}{c}{$\mu = 8$} & \multicolumn{1}{c}{$\mu = 16$} & \multicolumn{1}{c}{$\mu = 32$} & \multicolumn{1}{c}{$\mu = 64$} & \multicolumn{1}{c}{$\mu = 128$} \\
		\midrule
		\texttt{Image Size (kB)} &143 &132 &129 &126 &125 \\
		\texttt{PSNR (dB)} &20.6457 &25.3002 &29.3269  &30.8296 &35.4472\\ 
		\texttt{CS} &1.0699 &1.1590 &1.1860 &1.2142 &1.224 \\ 
		\bottomrule          
	\end{tabular}
\end{table}
\clearpage

\end{document}